\documentclass[10pt,twocolumn,letterpaper]{article}

\usepackage[pagenumbers]{cvpr} 
\usepackage[accsupp]{axessibility}

\usepackage[dvipsnames]{xcolor}

\definecolor{cvprblue}{rgb}{0.21,0.49,0.74}
\usepackage[pagebackref,breaklinks,colorlinks,citecolor=cvprblue]{hyperref}

\def\paperID{11205} 
\def\confName{CVPR}
\def\confYear{2024}

\title{Semantics-aware Motion Retargeting with Vision-Language Models}

\newcommand*{\affaddr}[1]{#1} 
\newcommand*{\affmark}[1][*]{\textsuperscript{#1}}

\author{%
Haodong Zhang\affmark[1]$^*$\quad Zhike Chen\affmark[1]$^*$\quad Haocheng Xu\affmark[1]\quad Lei Hao\affmark[2]\quad Xiaofei Wu\affmark[2]\quad\\ Songcen Xu\affmark[2]\quad Zhensong Zhang\affmark[2]\quad Yue Wang\affmark[1]\quad Rong Xiong\affmark[1]$^\dag$\\
\affaddr{\affmark[1]Zhejiang University}\quad
\affaddr{\affmark[2]Huawei Noah’s Ark Lab}\\
}

\begin{document}
\maketitle
\def\thefootnote{*}\footnotetext{These authors contributed equally to this work}
\def\thefootnote{\dag}\footnotetext{Corresponding author: rxiong@zju.edu.cn}
\begin{abstract}
Capturing and preserving motion semantics is essential to motion retargeting between animation characters. However, most of the previous works neglect the semantic information or rely on human-designed joint-level representations. Here, we present a novel \textbf{S}emantics-aware \textbf{M}otion re\textbf{T}argeting (SMT) method with the advantage of vision-language models to extract and maintain meaningful motion semantics. We utilize a differentiable module to render 3D motions. Then the high-level motion semantics are incorporated into the motion retargeting process by feeding the vision-language model with the rendered images and aligning the extracted semantic embeddings. To ensure the preservation of fine-grained motion details and high-level semantics, we adopt a two-stage pipeline consisting of skeleton-aware pre-training and fine-tuning with semantics and geometry constraints. Experimental results show the effectiveness of the proposed method in producing high-quality motion retargeting results while accurately preserving motion semantics. Project page can be found at \url{ https://sites.google.com/view/smtnet}.
\end{abstract}
\section{Introduction}
\label{sec:intro}

3D animation characters have extensive application in animation production, virtual reality, and various other domains. These characters are animated using motion data, resulting in lifelike and immersive animations. Nevertheless, acquiring motion data for each character can be a costly endeavor. Therefore, the ability to retarget existing motion data for new characters holds immense importance. The goal of motion retargeting is to transfer existing motion data to new characters following motion feature extraction and integration processes, which ensure the preservation of the original motion's characteristics.

Semantics encompasses the meaningful and contextually relevant information conveyed in motion and plays a critical role in ensuring the realism and vividness of the animation characters.
Preservation of motion semantics can enhance the efficiency of motion retargeting by reducing the need for time-consuming manual adjustments and refinements.
However, previous methods \cite{villegas2018neural,lim2019pmnet,aberman2020skeleton} are mainly based on retargeting of joint positions and make less use of the extraction of semantic information. They focus on trajectory-level motion retargeting with few attention to motion semantics. Consequently, this leads to a significant loss of motion semantics and necessitates the labor-intensive intervention of animation artists for manual trajectory adjustments. Recent advancements have introduced self-contacts \cite{villegas2021contact} and joint distance matrices \cite{zhang2023skinned} as the representation of motion semantics. Nevertheless, self-contacts are not applicable to non-contact semantics and require intricate vertex correspondence. The human-designed joint distance matrices primarily focus on joint relative relationships and still lack consideration of high-level semantic information.

\begin{figure}[t]
\centering
\includegraphics[width=\linewidth]{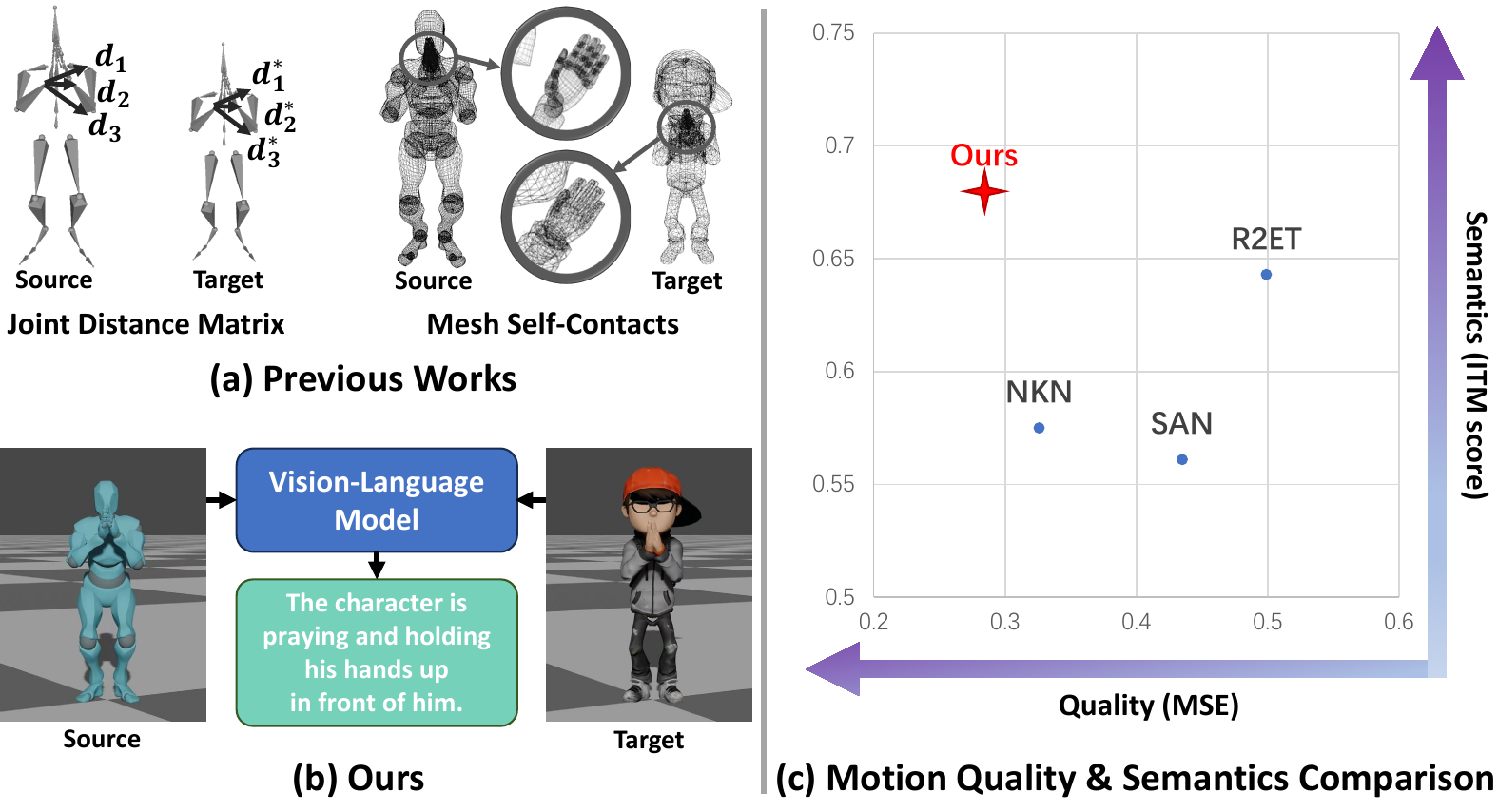}
\caption{
Comparison with previous motion retargeting methods. (a) Previous works rely on human-designed joint distance matrix \cite{zhang2023skinned} or self-contacts between mesh vertices \cite{villegas2021contact} to ensure semantics preservation. (b) Ours work enforces human-level motion semantics consistency with the extensive knowledge of vision-language models. (c) Comparison of motion quality and semantics preservation on the Mixamo dataset \cite{Mixamo}. Our method achieves the best motion quality and semantics consistency.
}
\label{introduction}
\vspace{-0.3cm}
\end{figure}

To address the intricate task of capturing and preserving motion semantics, we introduce a new perspective: the most general and comprehensive form of motion semantics is human-level natural language, reflecting the user's intuitive understanding of motion. However, the main challenge of human-level motion semantics representation lies in the scarcity of labelled data. It is difficult and expensive to label sufficient semantic textual descriptions for motion data.

In this paper, we introduce the incorporation of robust, state-of-the-art vision-language models to provide semantic guidance to the motion retargeting network. In the absence of labelled semantic data, we leverage the capabilities of a vision-language model to serve as a semantic supervisor in an unsupervised manner, which can extract motion semantics in a more intuitive way, as illustrated in Fig.~\ref{introduction}. This approach offers a solution to the challenge of the limited availability of labelled semantic datasets for motion retargeting.
To establish a connection between the vision-language model and motion semantics extraction, we employ the differentiable skinning and rendering modules to translate 3D motions into image sequences. Subsequently, we adopt visual question answering with guiding questions to inquire about the most relevant motion semantics from the vision-language model.

To guarantee the preservation of motion semantics during motion retargeting, we introduce a semantics consistency loss that enforces the semantic embeddings of the retargeted motion to closely align with those of the source motion. For dense semantic supervision and computational efficiency, we utilize latent features extracted by the vision-language model as the semantic embeddings instead of textual descriptions. To alleviate the non-linearity of the semantics consistency loss, we introduce a two-stage training approach. We categorize motion information into two distinct levels: the skeletal level and the semantic level. Our approach involves pre-training the motion retargeting network at the skeletal level, which is then further refined and fine-tuned at the semantic level with the power of vision-language models.
To the best of our knowledge, we are the first to leverage the extensive capability of vision-language models for the task of semantics-aware motion retargeting.




To summarize, the contributions of our work include:

\begin{itemize}

    \item We introduce an innovative framework that leverages the expertise of vision-language models as a semantic supervisor to tackle the challenge of limited labelled semantic data for the task of motion retargeting.
    

    \item We propose to use differentiable skinning and rendering to translate from the motion domain to the image domain and perform guiding visual question answering to obtain human-level semantic representation.
    
    \item We design a semantics consistency loss to maintain motion semantics and introduce an effective two-stage training pipeline consisting of pre-training at the skeletal level and fine-tuning at the semantic level.
    


    \item Our model achieves state-of-the-art performance in the challenging task of semantics-aware motion retargeting, delivering exceptional performance marked by high-quality motion and superior semantics consistency.

\end{itemize}
\section{Related Works}

\textbf{Optimization-based Motion Retargeting.}
Motion retargeting is a technique to adapt existing motion data from a source character to a target character with different bone proportions, mesh skins, and skeletal structures.
Early works formulate motion retargetting as a constrained optimization problem \cite{gleicher1998retargetting,popovic1999physically,lee1999hierarchical,choi2000online}.
Gleicher \etal \cite{gleicher1998retargetting} introduced a motion retargeting method, which identifies motion features as constraints and computes an adapted motion using a space-time constraint solver to preserve the desirable qualities.
Lee \etal \cite{lee1999hierarchical} proposed a method to adapt existing motion of a human-like character to have the desired features with specified constraints and combined a hierarchical curve fitting technique with inverse kinematics.
Nonetheless, these methods necessitate the tedious and time-consuming process of formulating human-designed constraints for specific motion sequences.

\noindent\textbf{Learning-based Motion Retargeting.}
With the rise of deep learning, researchers have been developing learning-based motion retargeting methods in recent years \cite{villegas2018neural,lim2019pmnet,aberman2020skeleton,villegas2021contact,hu2023pose,zhang2023skinned}.
Villegas \etal \cite{villegas2018neural} presented a recurrent neural network architecture, which incorporates a forward kinematics layer and cycle consistency loss for unsupervised motion retargetting.
Aberman \etal \cite{aberman2020skeleton} designed a skeleton-aware network with differentiable convolution, pooling, and unpooling operators to transform various homeomorphic skeletons into a primary skeleton for cross-structural motion retargeting.
However, these methods tend to concentrate on trajectory-level motion retargeting with limited consideration for motion semantics, which often results in a notable loss of motion semantics and increase the heavy burden of manual adjustments to the trajectories.
To address these problems, Zhang \etal \cite{zhang2023skinned} presented a residual retargeting network that uses a skeleton-aware module to preserve motion semantics and a shape-aware module to reduce interpenetration and contact missing. 
While this method successfully preserves joint relative relationships, it still falls short in addressing high-level motion semantics.

\begin{figure*}[htbp]
\centering
\includegraphics[width=0.95\linewidth]{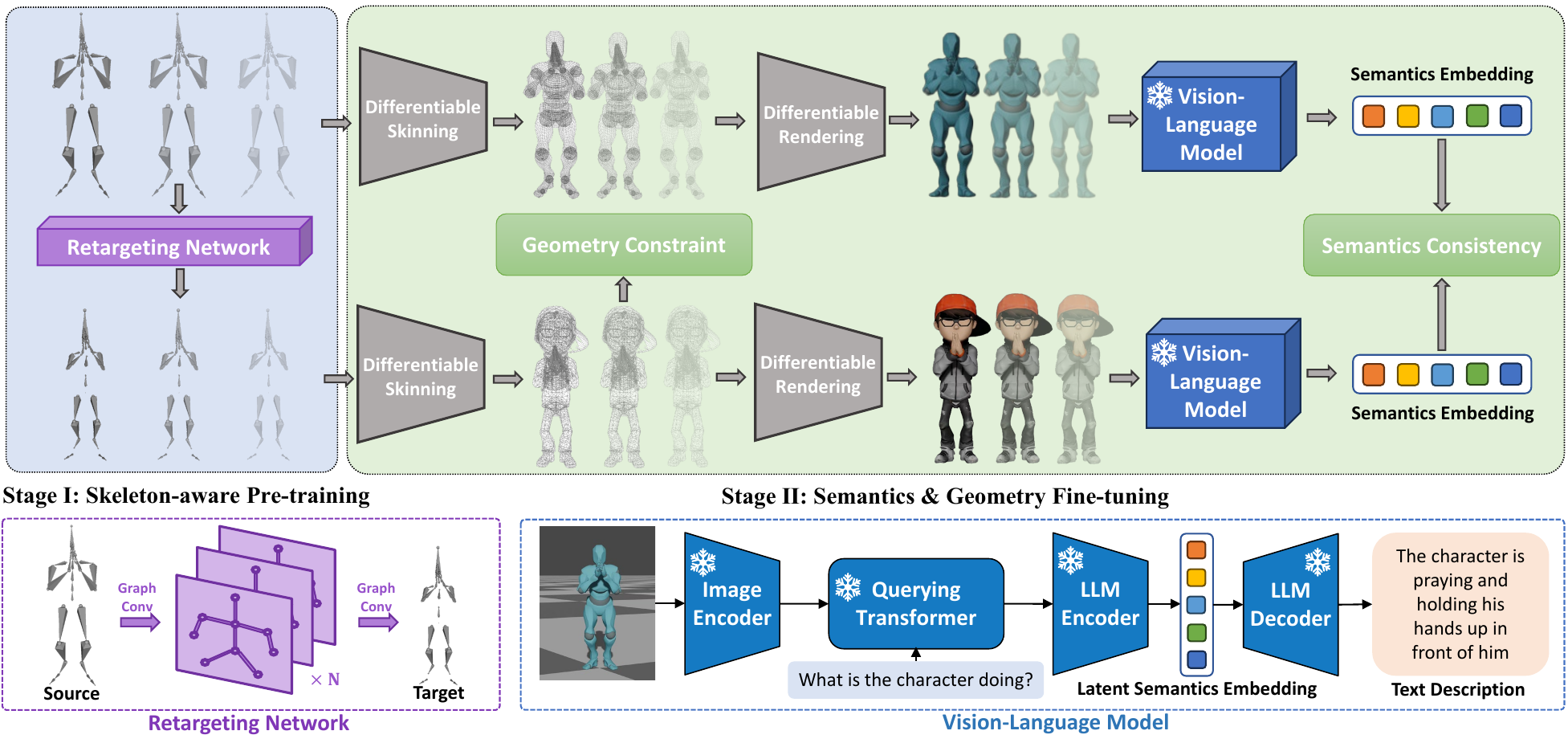}
\caption{
\textbf{Model Architecture.} Our semantics-aware motion retargeting framework employs a two-stage pipeline. Initially, the retargeting network consisting of multiple spatial-temporal graph convolution layers is trained at the skeletal level to establish a base model. Subsequently, this model undergoes further refinement and fine-tuning at the semantic level by the alignment of latent semantic embeddings of the source and target, leveraging the extensive knowledge of vision-language models. The latent semantic embedding is extracted by guiding visual question answering. Additionally, the geometry constraints are also enforced during fine-tuning to avoid interpenetration.
}
\label{framework}
\vspace{-0.3cm}
\end{figure*}

\noindent\textbf{Vision-Language Models.}
Vision-language models have empowered various vision-language tasks, including visual question answering and image captioning.
Tevet \etal \cite{tevet2022motionclip} introduced a human motion generation model that aligns the latent space with that of the Contrastive Language-Image Pre-training (CLIP) model.
Li \etal \cite{li2023blip} proposed a pre-training strategy from off-the-shelf frozen pre-trained image encoders and frozen large language models for vision-to-language generative learning.
Zhu \etal \cite{zhu2023minigpt} presented a vision-language model, which uses one projection layer to align a frozen visual encoder with a frozen advanced large language models (LLM).
However, these efforts primarily concentrate on vision-language tasks, leaving the question of how to effectively employ vision-language models to guide motion retargeting as an open and unexplored area.

\noindent\textbf{Human motion synthesis.}
Human motion synthesis is a domain related to motion retargeting, which aims to 
synthesize realistic and lifelike human motions from random noise or other inputs with generative networks.
Guo \etal \cite{guo2020action2motion} proposed to generate human motion sequences based on action type.
Guo \etal \cite{guo2022generating} presented a temporal variational autoencoder to synthesize human motions from text input.
Tevet \etal \cite{tevet2022human} introduced a diffusion-based generative model for human motion generation.
As comparison, we focus on the task of motion retargeting, where existing motion data is transferred from a source character to a target character.
\section{Method}

\subsection{Overview}

We present a novel semantic-aware motion retargeting method, as illustrated in Fig~\ref{framework}. In contrast to previous methods that neglect motion semantics \cite{villegas2018neural,lim2019pmnet,aberman2020skeleton} or rely on human-designed joint-level representations \cite{zhang2023skinned}, our approach integrates natural language descriptions from vision-language models to offer an explicit and comprehensive semantic representation of character motions, thereby maintaining the preservation of semantic consistency.

\noindent\textbf{Task definition.}
Given a source motion sequence, consisting of the skeleton motion and its associated skinning geometry, as well as a target character in the reference pose (e.g., T-pose), the objective of motion retargeting is to generate the target motion while preserving crucial motion characteristics, such as joint trajectory similarity and motion semantics, and satisfying geometry constraints.

\noindent\textbf{Graph representation.}
The skeleton motion sequence can be modelled as a sequence of graphs according to the skeleton hierarchy where each node corresponds to a joint and each edge represents a directed connection between joints. Assume that the motion sequence has $T$ frames in total and the animation characters have $N$ nodes and $M$ edges. In our approach, we consider motion data as node features $\mathbf{Q} \in \mathbb{R}^{T\times N\times 9}$, which encompass the 6D joint rotation representation \cite{zhou2019continuity} and 3D joint positions. Additionally, we utilize skeleton hierarchy information as edge features $\mathbf{E} \in \mathbb{R}^{M\times 3}$, which consists of the 3D position offset between each joint and its parent joint.

\noindent\textbf{Two-stage training.}
The motion of animation characters can be divided into skeletal movements and skinned movements, represented by skeletal joints and skinned vertices respectively. The skinned movements can be derived from the skeletal movements through the linear blend skinning algorithm \cite{lewis2000pose}. Therefore, motion retargeting at the skeletal level can effectively downscale the data and reduce the complexity of the problem. However, this simplification process can lead to the loss of motion semantics and violations of geometry constraints. To address these issues, we employ a two-stage pipeline. Initially, we pre-train a skeleton-aware network to ensure a general initialization for motion retargeting without considering motion semantics and geometry constraints. Subsequently, we fine-tune the pre-trained network for each source-target character pair with the vision-language model to maintain semantic consistency and enforce geometry constraints to prevent interpenetrations.

\subsection{Skeleton-aware Pre-training}

\noindent\textbf{Retargeting network.}
We propose a retargeting network consisting of a graph motion encoder and a graph motion decoder for motion retargeting. The motion encoder $\mathcal{F}_\theta$ encodes the motion data $\mathbf{Q}_A$ of the source character A into the latent motion embedding $\mathbf{Z}_A$. Then, the motion decoder $\mathcal{F}_\phi$ generates the joint angles $\mathbf{Q}_B$ of the target character B based on the latent features. Both the motion encoder and decoder are composed of multiple graph convolutions. More details are available in the supplementary materials.

\begin{equation}
\begin{aligned}
    \mathbf{Z}_A = \mathcal{F}_\theta(\mathbf{Q}_A, \mathbf{E}_A) \\
    \mathbf{Q}_B=\mathcal{F}_\phi(\mathbf{Z}_A, \mathbf{E}_B)
\end{aligned}
\end{equation}

In the first phase, we train the motion encoder and decoder at the skeletal level to establish a robust initialization for motion retargeting. Following the unsupervised learning setting in \cite{villegas2018neural}, we train the network with the reconstruction loss, cycle consistency loss, adversarial loss, and joint relationship loss. The overall objective function for skeleton-aware pre-training is defined as follows:

\begin{equation}
\label{objective}
     \mathcal{L}_{skel}=\lambda_r  \mathcal{L}_{rec}+\lambda_c \mathcal{L}_{cyc} + \lambda_a \mathcal{L}_{adv} + \lambda_j \mathcal{L}_{jdm}
\end{equation}

\noindent\textbf{The reconstruction loss} $\mathcal{L}_{rec}$ encourages the retargeted motion to match the source motion when the target character is the same as the source character. Let $\mathbf{Q}_{A,t}$ be the motion data of source character A at frame $t$, and $\hat{\mathbf{Q}}_{A,t}^{rec}$ be the reconstructed motion. Then $\mathcal{L}_{rec}$ is defined as:
\begin{equation}
\mathcal{L}_{rec}=\sum_t\left|\left|\hat{\mathbf{Q}}_{A,t}^{rec}-\mathbf{Q}_{A,t}\right|\right|^2_2
\end{equation}

\noindent\textbf{The cycle consistency loss} $\mathcal{L}_{cyc}$ promotes the consistency of retargeted motion from the source character A to the target character B and then back to the source character A, ensuring it remains in line with the original motion. Let $\hat{\mathbf{Q}}_{A,t}^{cyc}$ represent the retargeted motion.Then $\mathcal{L}_{cyc}$ is defined as:

\begin{equation}
    \mathcal{L}_{cyc}=\sum_t\left|\left|\hat{\mathbf{Q}}_{A,t}^{cyc}-\mathbf{Q}_{A,t}\right|\right|^2_2
\end{equation}

\noindent\textbf{The adversarial loss} $\mathcal{L}_{adv}$ is calculated by a discriminator network, which utilizes the unpaired data of the target character to learn how to distinguish whether the motions are real or fake. Let $\mathcal{F}_\gamma$ be the discriminator network, and $\mathbf{Q}_{B,t}$ be the retargeted motion at frame $t$. Then it is defined as:
\begin{equation}
    \mathcal{L}_{adv}=\sum_t\log\left(1-\mathcal{F}_\gamma\left(\mathbf{Q}_{B,t}\right)\right)
\end{equation}


\noindent\textbf{The joint relationship loss} $\mathcal{L}_{jdm}$ is calculated by the joint distance matrix (JDM) $\mathbf{D} \in \mathbb{R}^{N\times N}$, which represents the relative positional relationships of the joints. The element $d_{i,j}$ of $\mathbf{D}$ represents the Euclidean distance between joint $i$ and joint $j$. We extract the joint distance matrix from the target character and compare it with the source character. Then $\mathcal{L}_{jdm}$ is defined as:


\begin{equation}
  \mathcal{L}_{jdm} = \sum_t\left|\left|\eta(\mathbf{D}_{A,t}) - \eta(\mathbf{D}_{B,t})\right|\right|^2_2
  \label{eq:important}
\end{equation}
where $\eta(.)$ is an $\mathbf{L1}$ normalization performed on each row of the distance matrix. This normalization operation eliminates the difference in bone length to some extent.

\subsection{Semantics \& Geometry Fine-tuning}
In the second phase, we fine-tune the pre-trained retargeting network for each source-target character pair to preserve motion semantics and satisfy geometry constraints. The motion semantics is maintained by the semantics consistency loss, which aligns the semantic embeddings extracted from a vision-language model for both the source and target. Additionally, the geometry constraint is satisfied by minimizing the interpenetration loss. The overall objective function for fine-tuning is outlined as follows:

\begin{equation}
     \mathcal{L}_{fine}=\lambda_s \mathcal{L}_{sem} + \lambda_p  \mathcal{L}_{pen}
\end{equation}

\noindent\textbf{Differentiable skinning \& rendering.} To make the fine-tuning process differentiable for gradient back-propagation, we first use the differentiable linear blend skinning algorithm \cite{lewis2000pose}, denoted as $\mathcal{F}_{lbs}$, to transform the target joint angles $\mathbf{Q}_B$ into skinned motions $\mathbf{V}_B$, represented by 3D mesh vertices. Subsequently, we employ the differentiable projection function $\mathcal{F}_{proj}$ as introduced in \cite{liu2019soft} to convert the skinned motions into 2D images $\mathbf{I}_B$. A limitation for the differentiable rendering process is that when projecting the 3D skinned mesh onto 2D images, the depth information is lost. To obtain a comprehensive semantic representation of the motion, we render the character from multiple perspectives and then combine the extracted features, following the Non-rigid Shape Fitting task in \cite{liu2019soft}.

\begin{equation}
\begin{aligned}
    \mathbf{I}_A = \mathcal{F}_{proj}(\mathcal{F}_{lbs}(\mathbf{Q}_A)) \\
    \mathbf{I}_B = \mathcal{F}_{proj}(\mathcal{F}_{lbs}(\mathbf{Q}_B))
\end{aligned}
\end{equation}

\noindent\textbf{Frozen vision-language model.}
To obtain an explicit and reliable semantic feature of the motion, we employ a frozen vision-language model as our semantic supervisor. Current 3D vision-language datasets \cite{zhu2023vista,azuma_2022_CVPR} mainly focus on the occupation or the segmentation of the object in a spatial scene like rooms, and thus the state-of-the-art 3D vision-language models \cite{zhu2023vista} lack prior knowledge relevant to animation characters. In contrast, 2D vision-language models achieve better results in semantic tasks, such as image captioning, visual question answering and image-text retrieval, and provides cleaner and richer semantics \cite{yang2021sat}. Therefore, we utilize a frozen 2D vision-language model to extract latent embeddings of motion semantics.
The frozen 2D vision-language model employed in our work is BLIP-2 \cite{li2023blip2}, which incorporates a lightweight querying transformer as a bridge between the off-the-shelf frozen pre-trained image encoder and the frozen large language model.


\begin{figure}[t]
\centering
\includegraphics[width=\linewidth]{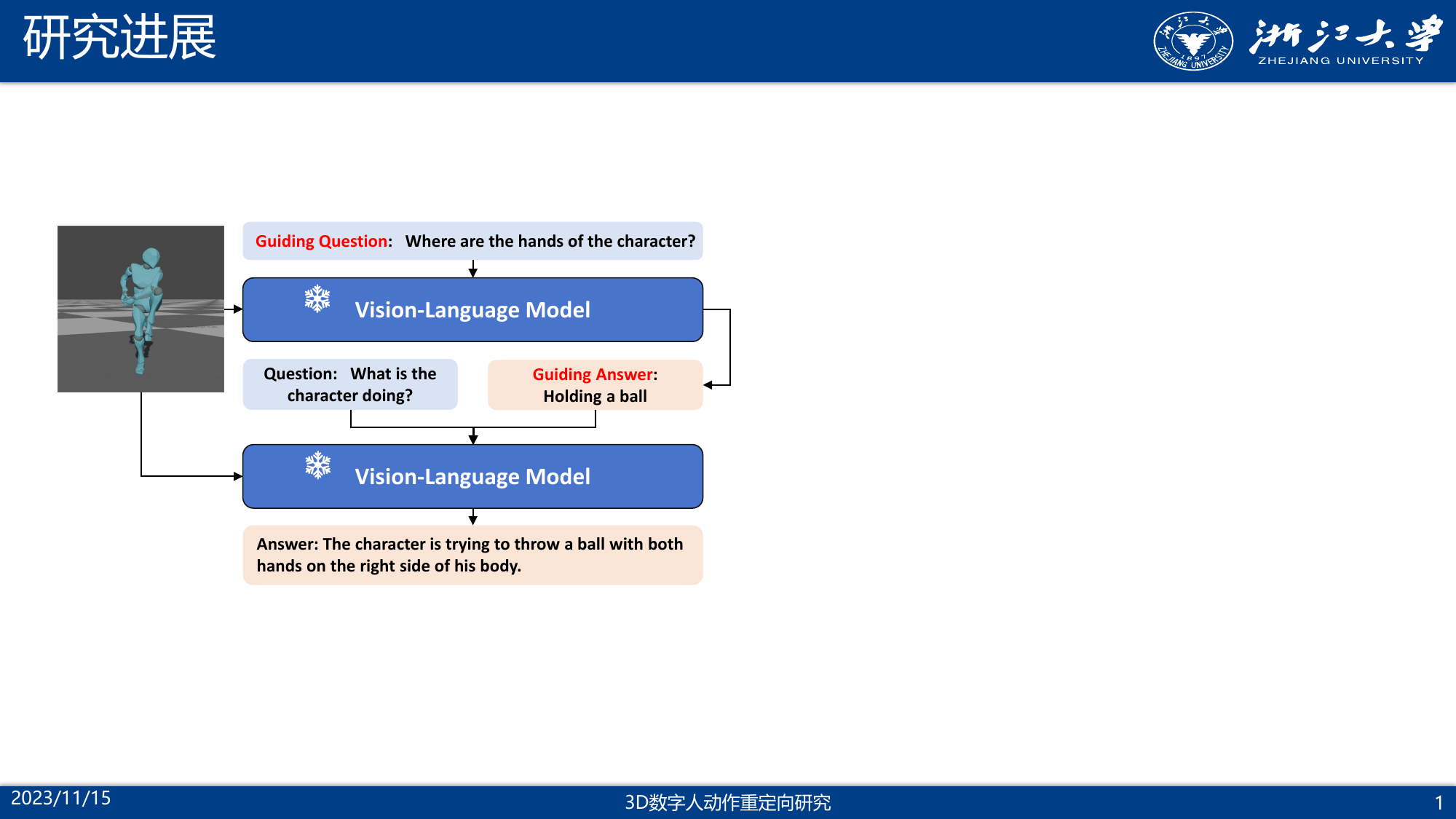}
\caption{
An example of guiding visual question answering.
}
\label{fig:-questionvlm}
\vspace{-0.3cm}
\end{figure}

\noindent\textbf{Prompt design.}
Since the vision-language model has the capability to extract rich information from images, it is possible that the extracted features might contain redundant details, such as the appearance of the character. To guide the vision-language model to obtain semantic embedding relevant to character motions, we adopt a guiding visual question answering approach for motion semantics extraction, as depicted in Fig.~\ref{fig:-questionvlm}. We believe that there is a strong correlation between motion semantics and hand movements. To acquire a more comprehensive description of the motion, we initially provide a guiding question to BLIP-2: \textit{``Where are the hands of the character?"}. Subsequently, we introduce a new question and combine it with the first answer as the input to BLIP-2: \textit{``[The answers to the first question generated by the vision-language model] What is the character in the image doing?"}. For more details, please refer to the supplementary materials.





\noindent\textbf{Latent semantic embedding.} We opt to align the latent semantic embeddings of the source and target generated by the vision-language model rather than relying on textual descriptions, specifically leveraging the encoder output of the large language model. This approach enables us to acquire a more accurate and denser representation, while also mitigating computational costs and the non-linearity of the training objective caused by the large number of parameters of the vision-language model. Let $\mathbf{E}_A$ and $\mathbf{E}_B$ be the latent semantic embeddings of the source and target motions, $\mathcal{F}_\omega$ be the frozen pre-trained image encoder, $\mathcal{F}_\sigma$ be the frozen querying transformer, $\mathcal{F}_\psi$ be the encoder of the frozen large language model, and $context$ be the question.

\begin{equation}
\begin{aligned}
    \mathbf{E}_A = \mathcal{F}_\psi(\mathcal{F}_\sigma(\mathcal{F}_\omega(\mathbf{I}_A), context)) \\
    \mathbf{E}_B = \mathcal{F}_\psi(\mathcal{F}_\sigma(\mathcal{F}_\omega(\mathbf{I}_B), context))
\end{aligned}    
\end{equation}

\noindent\textbf{Fine-tuning with semantics consistency.} As illustrated in Fig.~\ref{framework}, our approach aligns the latent semantic embeddings of both the source and target motions in an unsupervised manner, ensuring a high degree of semantic consistency in the retargeted results. The semantics consistency loss $\mathcal{L}_{sem}$ is calculated using the mean square error and it is defined as follows:

\begin{equation}
  \mathcal{L}_{sem} = \sum_t \left\| \mathbf{E}_{A,t} - \mathbf{E}_{B,t} \right\|_2^2
\end{equation}

\noindent\textbf{Fine-tuning with geometry constraints.}
From our observations, most interpenetration problems occur between the limbs and the main body. To address this, we incorporate the signed distance field between the limb vertices and the body mesh as the interpenetration loss. First, we convert the skeleton motion output from the network into mesh vertices using the linear blend skinning method \cite{lewis2000pose}. Then, the interpenetration loss is defined as follows:

\begin{equation}
  \mathcal{L}_{pen} = \sum_t ReLU(-\Phi_{b,t}(\mathbf{V}_{l,t}))
\end{equation}

\noindent where $\Phi_b$ indicates the signed distance field function, $\mathbf{V}_l$ is the vertices of the limbs. If the vertex locates inside the body, the value of the function is less than zero. Therefore, we use the $ReLU$ function to penalize the inner vertices.
\section{Experiments}

\subsection{Settings}

\noindent\textbf{Datasets.}
We train and evaluate our method on the Mixamo dataset \cite{Mixamo}, an extensive repository of animations performed by various 3D virtual characters with distinct skeletons and geometry shapes.
The training set we use to pre-train our skeleton aware module is the same as that used in \cite{aberman2020skeleton}, which contains 1646 motions performed by 7 characters.
It's important to note that the Mixamo dataset does not provide clean ground truth data, since many of the motion sequences suffer from interpenetration issues and semantic information loss.
To mitigate this, we have carefully selected a subset of motion sequences that are both semantically clean and free of interpenetration issues for fine-tuning and testing.
Our fine-tuning process involves retargeting 15 clean motions including 3127 frames, originally performed by 3 source characters, namely ``Y Bot", ``X Bot", and ``Ortiz", onto 3 target characters, including ``Aj", ``Kaya", and ``Mousey".
Then we evaluate the performance of our model on the task of retargeting 30 additional motions that are previously unseen in the training set and fine-tuning sets.
More details could be found in the supplementary materials.

\noindent\textbf{Implementation details.}
The hyper-parameters $\lambda_r$, $\lambda_c$, $\lambda_a$, $\lambda_j$, $\lambda_p$, $\lambda_s$ for pre-training and fine-tuning loss functions are set to $10.0, 1.0, 0.1, 1.0, 1.0, 0.1$.
For semantics fine-tuning, we use BLIP-2 \cite{li2023blip2} with pre-trained FlanT5-XXL \cite{chung2022scaling} large language model. To extract the semantic representation of the motion, we render animation from three perspectives, including the front view, left view and right view. The fine-tuning process takes 25 epochs with 5 clean motion sequences of the source character for each target character. During pre-training and fine-tuning, we use an Adam optimizer to optimize the retargeting network. Please refer to the supplementary materials for more details.

\noindent\textbf{Evaluation metrics.}
We evaluate the performance of our method across three key dimensions: skeleton, geometry, and semantics. At the skeletal level, we measure the Mean Square Error (MSE) between retargeted joint positions and the ground truth provided by Mixamo, analyzing both the global and the local joint positions. At the geometric level, we evaluate the interpenetration percentage (PEN). At the semantic level, we utilize the Image-Text Matching (ITM) score, Fr\'echet inception distance (FID) and semantics consistency loss (SCL) as metrics. The ITM score quantifies the visual-semantic similarity between the source textual description and the rendered retargeted motion. FID is calculated between the semantic embedding distribution of retargeted motion and source motion. More details are provided in the supplementary materials.

\begin{figure*}[htbp]
\centering
\includegraphics[width=\linewidth]{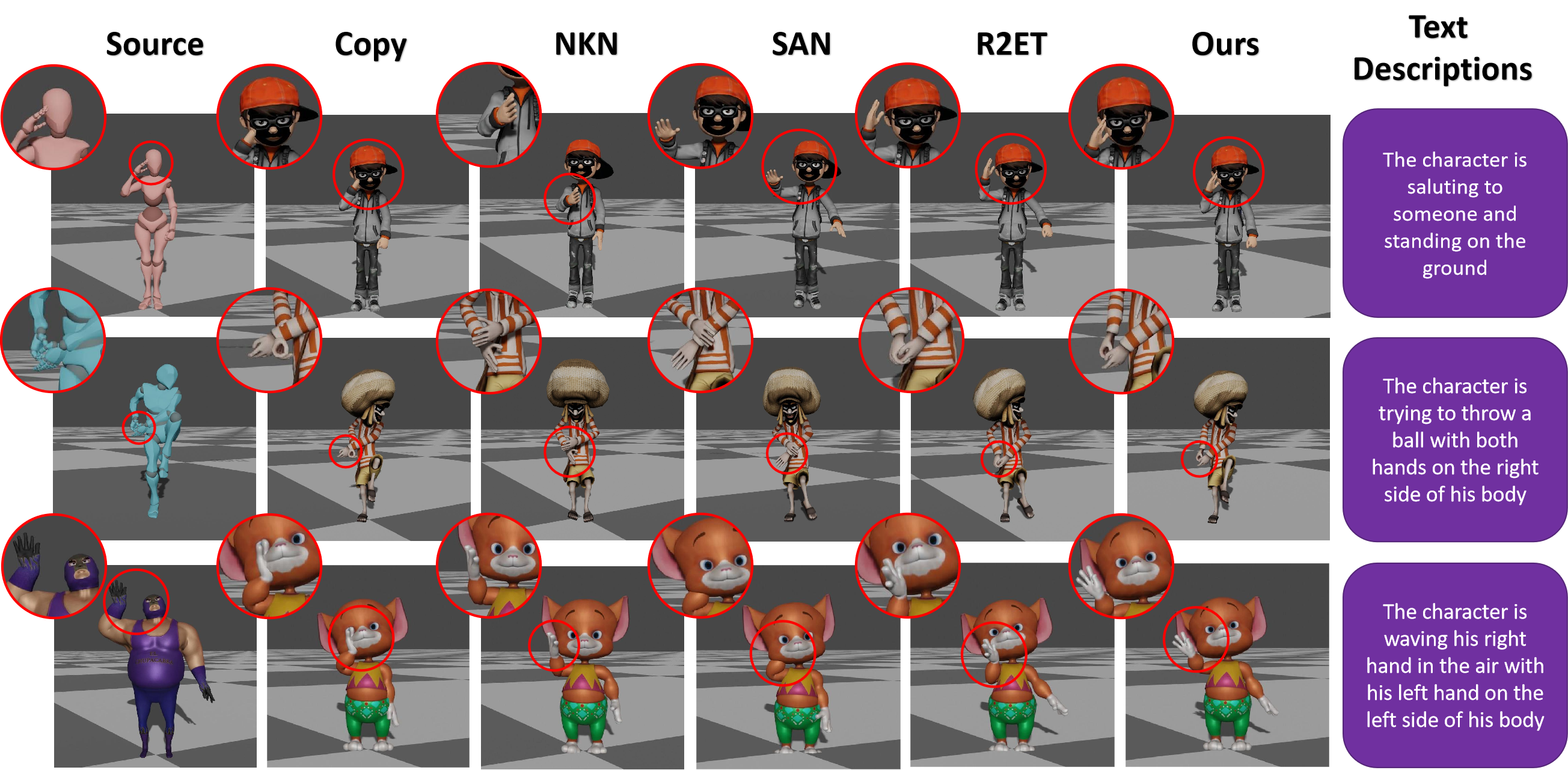}
\caption{
Qualitative comparision. The results demonstrate that our method can effectively preserve semantics while the baseline methods suffer from interpenetration or semantic information loss. From the first column to the last column are the source motion, the Copy strategy, NKN \cite{villegas2018neural}, SAN \cite{aberman2020skeleton}, R2ET \cite{zhang2023skinned}, our method and text descriptions, respectively.
}
\label{fig:comparision}
\vspace{-0.4cm}
\end{figure*}

\subsection{Comparison with State of the Arts}
\noindent\textbf{Quantitative.}
In this section, we conduct a comparative analysis of our method against the state-of-the-art approaches as illustrated in Tab.~\ref{tab:compare}.
The baseline methods include R2ET \cite{zhang2023skinned}, SAN \cite{aberman2020skeleton}, NKN \cite{villegas2018neural} and the Copy strategy.
The Copy strategy achieves the lowest local MSE because the ground truth data in the Mixamo dataset are not entirely clean, and many of them are generated by copying rotations. As a result, this strategy comes at the cost of semantic loss and interpenetration issues.
SAN \cite{aberman2020skeleton} and NKN \cite{villegas2018neural} focus on skeleton-level motion features, which results in a high interpenetration rate and relatively low semantics preservation.
R2ET \cite{zhang2023skinned} treats motion semantics as the joint distance matrix and mesh distance field, which helps it obtain better motion semantics than SAN and Copy. Nevertheless, there is still a gap between the human-designed distance matrix and the human-level semantics.
Notably, our model exhibits the best interpenetration rate and semantics preservation among all methods, showcasing the capability of the proposed method in producing high-quality retargeted motions with semantics consistency.

\noindent\textbf{Qualitative.} In Fig.~\ref{fig:comparision}, we visualize the text descriptions of the motions and the qualitative comparison between the state-of-the-arts and our method. SAN \cite{aberman2020skeleton} and Copy neglect the preservation of semantics and have severe interpenetration. R2ET \cite{zhang2023skinned} utilizes joint distance matrix as semantics representation and fails to capture high-level semantic information. For example, the salute motion retargeted by R2ET \cite{zhang2023skinned} appears more like a hand-up motion. As a comparison, our method is able to successfully preserve high-level motion semantics leveraging the vision-language model. We observe that our approach reaches the best results among all methods, achieving more reliable semantics preservation and lower interpenetration rates. It suggests that with semantics and geometry fine-tuning, our method could effectively solve interpenetration issues together with semantics preservation.

\begin{table}
\centering
\resizebox{\columnwidth}{!}{%
\begin{tabular}{@{}lcccccc@{}}
\toprule
Method & MSE $\downarrow$ & MSE$^{lc}$ $\downarrow$ & Pen.\% $\downarrow$ & ITM $\uparrow$ & FID $\downarrow$ &SCL $\downarrow$\\
\midrule
Source & - & - & 4.43 & 0.796 &- &-\\
GT & - & - & 9.06 & 0.582 &26.99 & 1.331\\ 
\midrule
Copy & - & \textbf{0.005} & 9.03 & 0.581 &26.58 &1.327\\
NKN \cite{villegas2018neural} & 0.326 & 0.231& 8.71 & 0.575 &27.79 &1.414\\
SAN \cite{aberman2020skeleton} & 0.435 & 0.255& 9.74 & 0.561 &  28.33& 1.448\\
R2ET \cite{zhang2023skinned} & 0.499& 0.496& 7.62 & 0.643 & 5.469& 0.405\\
Ours & \textbf{0.284} & 0.229 & \textbf{3.50} & \textbf{0.680} & \textbf{0.436} &\textbf{0.143}\\
\bottomrule
\end{tabular}
}
\vspace{-0.2cm}
\caption{Quantitative comparison with the state-of-the-arts. MSE$^{lc}$ denotes the local MSE. ITM indicates the image-text matching score. FID is Fr\'echet inception distance of motion semantics. SCL is the semantics consistency loss.}
\label{tab:compare}
\vspace{0.2cm}

\resizebox{\columnwidth}{!}{%
\begin{tabular}{@{}lcccccc@{}}
\toprule
Method & MSE $\downarrow$ & MSE$^{lc}$ $\downarrow$ & Pen.\% $\downarrow$ & ITM $\uparrow$ & FID $\downarrow$ &SCL $\downarrow$\\
\midrule
SMT$_{tws}$ & \textbf{0.248} & \textbf{0.129}& 8.37 & 0.586&7.727 &0.769 \\
SMT$_{twf}$ &7.798 &7.083 &\textbf{0.44} &0.432& 56.53 & 13.29\\
SMT$_{twa}$ & 0.335&0.288&5.36&0.658& 2.826&0.266\\
SMT$_{fwp}$ &0.439 & 0.368 & 1.22 & 0.597& 7.241&0.583\\
SMT$_{fwi}$ &5.418 &4.576 &4.41 &0.552 &78.46 &18.96\\
SMT$_{fwq}$ &0.739 &0.517 &4.56 &0.668& 2.497& 0.191\\
SMT$_{Ours}$ &0.284 &0.229 &3.50 & \textbf{0.680} & \textbf{0.436} &\textbf{0.143} \\
\bottomrule
\end{tabular}
}
\vspace{-0.2cm}
\caption{Ablation study. SMT$_{tws}$ is the network trained with only skeleton-aware pre-training. SMT$_{twf}$ is the network trained with only semantics and geometry fine-tuning. SMT$_{twa}$ is the network trained in one stage. SMT$_{fwp}$ is the network fine-tuned with only the interpenetration loss. SMT$_{fwi}$ is the network fine-tuned with image features. SMT$_{fwq}$ is the network fine-tuned with the features of the querying transformer.}
\label{tab:ablation}
\vspace{-0.5cm}
\end{table}

\begin{figure}[t]
\centering
\vspace{1pt}
\includegraphics[width=0.9\linewidth]{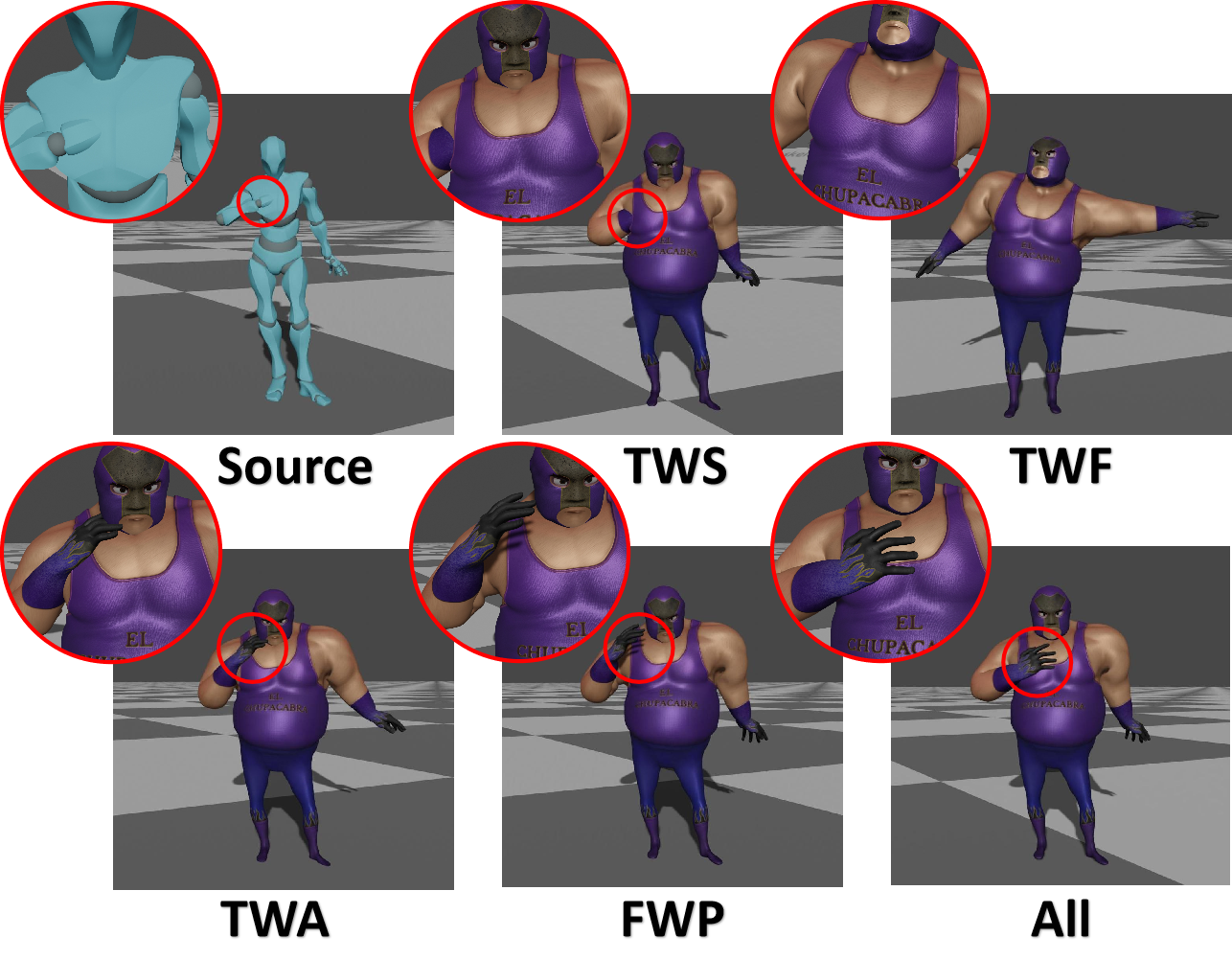}
\vspace{-0.3cm}
\caption{
The qualitative comparison of ablation study between the network without fine-tuning (TWS), the network trained with only semantics and geometry fine-tuning (TWF), the network trained with all loss functions (TWA), the network fine-tuned with only the interpenetration loss (FWP) and our full model (All).
}
\label{fig:ablation}
\vspace{-0.7cm}
\end{figure}

\subsection{Ablation Studies}

\noindent\textbf{Skeleton-aware pre-training.}
The proposed method can be divided into two stage: pre-training and fine-tuning.
To illustrate the importance of skeleton-aware pre-training,
we evaluate the network trained with only the semantics consistency loss and the interpenetration loss in Tab.~\ref{tab:ablation}, denoted as SMT$_{twf}$. The network trained without skeleton-aware pre-training performs worst in MSE and semantics preservation. A reasonable explanation is that the semantics consistency loss is highly non-linear, so it is important to pre-train the network at the skeletal level to provide better initial values. We also visualize qualitative results in Fig.~\ref{fig:ablation}.

\noindent\textbf{Semantics \& geometry fine-tuning.}
We also conduct ablation study to illustrate the importance of semantics and geometry fine-tuning in Tab.~\ref{tab:ablation}.
We first evaluate the performance of the skeleton-aware model without fine-tuning, denoted as SMT$_{tws}$. Though it reaches the best global position MSE, it suffers from interpenetration and semantic information loss because of the low-quality motion data provided by Mixamo.
We next evaluate the network fine-tuned with only the interpenetration loss, denoted as SMT$_{fwp}$. This version results in a significant boost in terms of penetration rate. However, the gradient of interpenetration loss is only relevant with the face normals of the geometry mesh without considering the semantic information conveyed in the motion. It indicates the importance of the semantic consistency loss that makes the network reach a better balance between interpenetration and semantics.
We also try to train the network with all loss functions in one stage, denoted as SMT$_{twa}$. However, it is challenging for the model to acquire general knowledge of interpenetration and semantics that is suitable for every character with limited data.
Therefore, training the model with skeleton-aware pre-training and fine-tuning it with semantics consistency and geometry constraints for each target character remains a more reasonable and data-efficient strategy.



\noindent\textbf{Latent semantic embedding.}
The vision-language model used for semantic extraction can be divided into three parts: the image encoder from CLIP \cite{radford2021learning}, the querying transformer and the large language model. In Tab.~\ref{tab:ablation}, we compare the feature outputted by the image encoder, the querying transformer and the encoder of the large language model, denoted as SMT$_{fwi}$, SMT$_{fwq}$, and SMT$_{Ours}$, respectively. The results show that the image feature performs worse since it is greatly affected by the appearance of the character. It indicates that with the help of the large language model, the semantic representation better focuses on the semantic meaning of the motion instead of the character's visual appearance. Therefore, the encoder output of the large language model is more suitable for semantic embedding. More details can be found in the supplementary materials.

\begin{figure}[t]
\centering
\includegraphics[width=\linewidth]{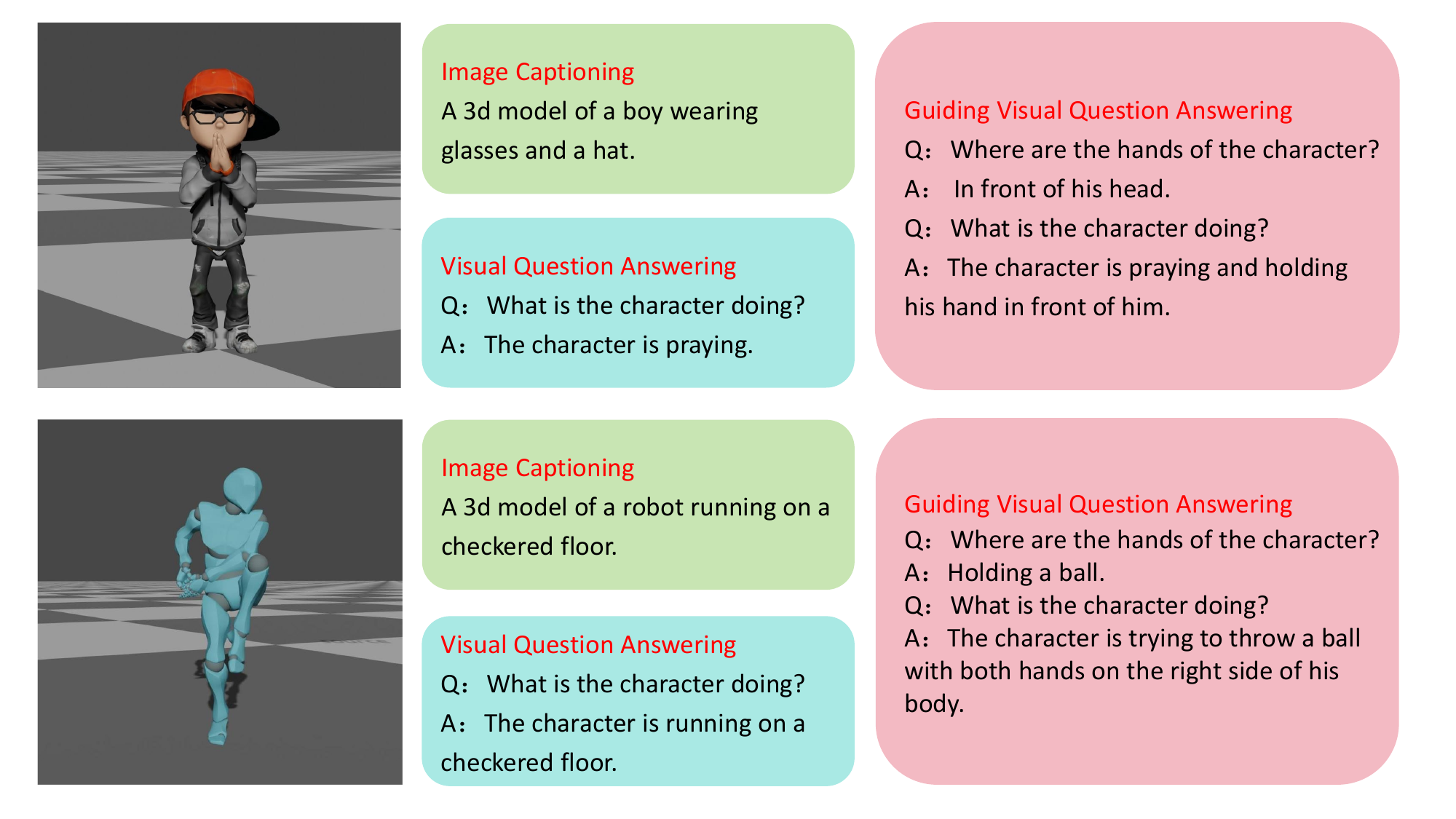}
\vspace{-0.5cm}
\caption{
Text descriptions generated by different ways. The guiding visual question answering yields more comprehensive results.
}
\label{fig:vqa}
\vspace{-0.2cm}
\end{figure}

\noindent\textbf{Prompt design.}
To validate the importance of guiding visual question answering, we compare the textual descriptions generated by visual question answering with and without guiding questions as well as image captioning. The results in Fig.~\ref{fig:vqa} indicate that using guiding questions for visual question answering yields the most comprehensive and reasonable text descriptions for motion semantics. Compared with image captioning that uses the vision-language model to generate text description directly from images, the answers from visual question answering task can be guided by the designed question to focus on motion semantics. 

\begin{table}
\centering
\begin{tabular}{@{}lccccc@{}}
\toprule
Method & Quality $\uparrow$ & Smoothness $\uparrow$ & Semantics $\uparrow$\\
\midrule
Copy &0.72& \textbf{0.86}& 0.71 \\
NKN \cite{villegas2018neural} &0.65 & 0.80& 0.66\\
SAN \cite{aberman2020skeleton} &0.69& 0.82& 0.67 \\
R2ET \cite{zhang2023skinned} &0.80& 0.61& 0.85 \\
Ours &\textbf{0.89}& 0.80& \textbf{0.92} \\
\bottomrule
\end{tabular}
\vspace{-0.2cm}
\caption{User study results. We collect 100 comparisons in three aspects. Our method gets highest scores in the overall quality as well as semantics preservation.}
\label{tab:user}
\vspace{-0.6cm}
\end{table}

\subsection{User Study}
We conduct a user study to evaluate the performance of our method against the baseline methods. Human subjects are given 12 videos. Each video includes one source skinned motion and five anonymous skinned results. The retargeted results are randomly placed. We ask subjects to rate the results out of 1.0 in three aspects: overall quality, motion smoothness and semantics preservation. We collect a total of 100 comparisons. During the evaluation, users are required to extract semantic meaning from the source motion themselves and then evaluate the preservation of retargeted motions. In general, more than 92\% of subjects prefer the retargeting results of our method.

\begin{figure}[t]
\centering
\includegraphics[width=0.9\linewidth]{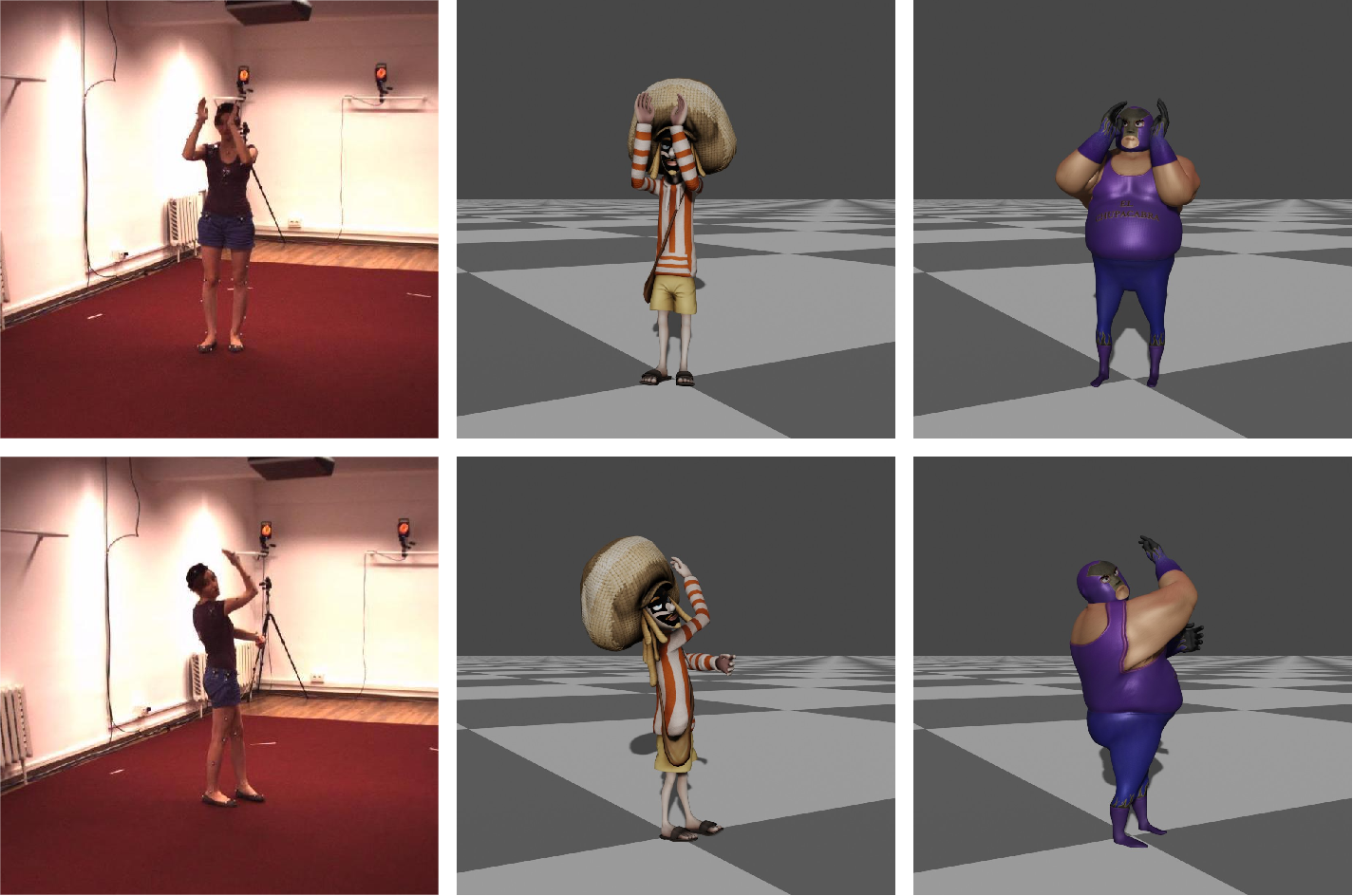}
\vspace{-0.2cm}
\caption{
We retarget from human motion clips in the human3.6M \cite{h36m_pami} dataset. The retargeted motions are free from interpenetration and preserve semantics well.
}
\label{fig:video}
\vspace{-0.6cm}
\end{figure}

\subsection{Retargeting Motion from Human Videos}

In this section, we evaluate our motion retargeting approach from human videos in the human3.6M \cite{h36m_pami} dataset. Video retargeting involves two stages: human pose estimation from video and motion retargeting. However, inaccuracies in estimating body postures may result in semantic information loss and thus accumulation of errors in the entire retargeting process. Therefore, we first get the estimated human pose from \cite{Moon_2022_CVPRW_Hand4Whole}. Then we utilize the vision-language model to extract the semantic embedding of the original video and calculate the semantic consistency loss to optimize the joint angles acquired from the retargeting process directly. In Fig.~\ref{fig:video}, we show our results of motion retargeting from human videos to Mixamo characters.

\section{Conclusions}

In this paper, we present a novel semantics-aware motion retargeting method that leverages the capabilities of vision-language models to extract semantic embeddings and facilitate the preservation of motion semantics. This approach offers a promising solution to the challenge of lacking labelled semantic data for motion. Our proposed method involves a two-stage process that integrates skeleton-level motion characteristics and semantics-level consistency along with geometry constraints. Experimental results demonstrate that our approach excels in generating high-quality retargeted motions with semantics consistency.

\noindent\textbf{Limitations.}
The main limitation is the performance of the vision-language model in extracting motion semantics. Without the support of motion semantic datasets of sufficient data size and quality, we rely on the model pre-trained on large image-text datasets. Although the model achieves some remarkable results in motion semantics extraction, there is still room for improvement. In addition, the projection of 3D motion into 2D images loses spatial information and affects the performance.

\noindent\textbf{Future work.}
Compared with 2D vision-language models, 3D vision-language models have the advantage of capturing spatial relationships directly.
Therefore, fine-tuning 3D vision-language models to make them more suitable for the task of motion semantics extraction is worth exploring in our future work.

\noindent\textbf{Acknowledgements.}
This work was supported by the National Nature Science Foundation of China under Grant 62173293.
\newpage
{
    \small
    \bibliographystyle{ieeenat_fullname}
    \bibliography{main}
}

\clearpage
\setcounter{page}{1}
\maketitlesupplementary
\renewcommand{\thesubsection}{\Alph{subsection}}

In this document, we provide the following supplementary content:
\begin{itemize}[topsep=5pt,parsep=5pt]
    \item Dataset Details
    \item Evaluation Metrics
    \item Implementation Details
    \item Semantic Attention Visualization
    \item Ablation Study
    \item Additional Results
\end{itemize}

\subsection{Dataset Details}

\textbf{Test set and fine-tuning set details.}
We download 45 source motions of source characters including Y Bot, X Bot and Ortiz for testing and fine-tuning. More details can be found in Tab.~\ref{tab:dataset}. As we focus on the preservation of semantics, we choose the motions that are free of interpenetration and have obvious semantics which can be accurately described by text descriptions. To compute the MSE metric, we download the corresponding ground truth data provided by Mixamo of target characters including Aj, Kaya and Mousey. The ground truth is mainly created by copying the rotations of corresponding joints from the source character thus suffering from interpenetration and semantic loss. In order to equip our model with abundant information about geometry skinning and motion semantics, we carefully choose motions that explore the entire movement space of each source character for fine-tuning the model. 

\subsection{Evaluation Metrics}

We evaluate the performance of our method across three key dimensions: skeleton, geometry, and semantics. At the skeletal level, we measure the Mean Square Error (MSE) between retargeted joint positions $\hat{\boldsymbol{P}}_B$ and ground truth $\boldsymbol{P}_B$ provided by Mixamo, normalized by the character height $h_B$. We compare both the global and the local joint positions. The local MSE is calculated when the root position is aligned with the ground truth.

\begin{equation}
    MSE = \frac{1}{h_B}\left|\left|\hat{\boldsymbol{P}}_B - \boldsymbol{P}_B\right|\right|^2_2
\end{equation}

At the geometric level, we evaluate the interpenetration percentage, which is calculated as the ratio of the number of penetrated vertices to the total number of vertices in each frame. A lower ratio indicates less interpenetration occurs.

\begin{equation}
    PEN = \frac{\textit{Number of penetrated vertices}}{\textit{Total number of vertices}}\times 100\%
\end{equation}

At the semantic level, we utilize the Image-Text Matching (ITM) score, Fr\'echet inception distance (FID) and semantics consistency loss as metrics to evaluate the semantics consistency. The task of Image-Text Matching \cite{li2019vsrn} is to measure the visual-semantic similarity between an image and a textual description via a two-class linear classifier $\mathcal{F}_c$ pre-trained in BLIP-2 \cite{li2023blip2}. To compute ITM, we first generate the textual description of the source motion with visual question answering and then compute the ITM score between the source textual description, denoted as $text$, and the rendered retargeted motion, denoted as $image$.


\begin{equation}
    ITM = \mathcal{F}_c(text, image)
\end{equation}

Fr\'echet inception distance (FID) is calculated between the semantic embedding distribution of retargeted motion and source motion. Let $\mathcal{N}(\mu_s,\Sigma_s)$ denotes the source distribution, while $\mathcal{N}(\mu_t,\Sigma_t)$ denotes the target distribution.
\begin{equation}
    FID = \left|\left|\mu_s - \mu_t\right|\right|^2_2 + tr(\Sigma_s + \Sigma_t - 2(\Sigma_s^{\frac{1}{2}}\Sigma_t\Sigma_s^{\frac{1}{2}})^{\frac{1}{2}})
\end{equation}

\subsection{Implementation Details}

\textbf{Training details.}
We use four NVIDIA 3090Ti (24*4GB) and the trainning process is divided into two stages.
For skeleton-aware pre-training, the learning rate is set as 0.0003, the number of training epoch is set as 80 and the batch size is 16.
For semantics fine-tuning, the learning rate is set to 0.0001 and batch size is 4. After 25 epoches, our model achieves state-of-the-art performance in motion retargeting and preserves the semantics of motion well. 
When fine-tuning our model with the interpenetration loss and the semantics consistency loss, we increase the weight of the interpenetration loss from 1.0 to 10.0 during the first 5 epochs. Because we observe that the performance of the vision-language model is unstable when there exists obvious interpenetration. And after 5 epoch, the weight goes back to 1.0.
The initial hyper-parameters $\lambda_r$, $\lambda_c$, $\lambda_a$, $\lambda_j$, $\lambda_p$, $\lambda_s$ for pre-training and fine-tuning loss functions are set to $10.0, 1.0, 0.1, 1.0, 1.0, 0.1$.
The vision language model we used is BLIP-2 \cite{li2023blip2} with pre-trained FlanT5-XXL \cite{chung2022scaling} large language model and large scale pre-trained vision transformer. In order to generate more comprehensive text for our prompt, we use beam search with a beam width
of 5. We also set the length-penalty to 1 which encourages longer answers.

\noindent\textbf{Network architecture.}
The motion encoder and decoder architectures consist of three layers of graph convolutions. The first two layers utilize spatial graph convolution, adopting a message-passing scheme to aggregate features from neighboring nodes as Eq.~\ref{gconv}. The last layer is the temporal graph convolution, which maintains the same number of channels. 
The motion encoder receives joint rotations and positions as input and encodes them into latent motion embeddings, expanding the channels from 9 to 16 and 32. Subsequently, the motion decoder takes these latent motion embeddings as input and outputs the target joint rotations, gradually reducing the channels from 32 to 16 and 6. Additionally, the root joint positions are generated using a two-layer MLP, starting with node features from the root joint and expanding channels to 16 and 3.

\begin{equation}
\label{gconv}{\boldsymbol{x}_{i}}^{'}=\boldsymbol{x}_i+\sum_{j\in{N(i)}}g(\boldsymbol{W}_{f}[\boldsymbol{x}_i,\boldsymbol{x}_j,\boldsymbol{e}_{j,i}]+\boldsymbol{b}_f)
\end{equation}

\noindent where $\boldsymbol{x}_i$ is the feature of node $i$, ${\boldsymbol{x}_{i}}^{'}$ is the updated feature of node $i$, $N(i)$ is the set of neighbor nodes of node $i$, and $\boldsymbol{e}_{j,i}$ is the edge feature from node $j$ to node $i$, $g$ is the LeakyReLU function, $\boldsymbol{W}_{f}$ and $\boldsymbol{b}_f$ are learnable parameters.

\begin{table}
\centering
\resizebox{\columnwidth}{!}{%
\begin{tabular}{@{}lcccccc@{}}
\toprule
Method & MSE $\downarrow$ & MSE$^{lc}$ $\downarrow$ & Pen.\% $\downarrow$ & ITM $\uparrow$ & FMD$\downarrow$ &SCL $\downarrow$ \\
\midrule
SMT$_{fwi}$ &5.418 &4.576 &4.41 &0.552& 78.46&18.96 \\
SMT$_{fwq}$ &0.739 &0.517 &4.56 &0.658&2.497 & 0.191 \\
SMT$_{Ours}$ &\textbf{0.284} &\textbf{0.229} &\textbf{3.50} & \textbf{0.680}&\textbf{0.436}&\textbf{0.143}\\
\bottomrule
\end{tabular}%
}
\caption{Ablation study on semantic embedding. We compare the performance of the model fine-tuned with the image feature from CLIP \cite{radford2021learning} as semantic embedding (SMT$_{fwi}$), the model fine-tuned with the features of the querying transformer as semantic embedding (SMT$_{fwq}$) and the model fine-tuned with the features of the larage language model encoder (SMT$_{Ours}$)}
\label{tab:compare1}
\vspace{-0.3cm}
\end{table}

\subsection{Semantic Attention Visualization}
\begin{figure}[t]
\centering
\vspace{6pt}
\includegraphics[width=\linewidth]{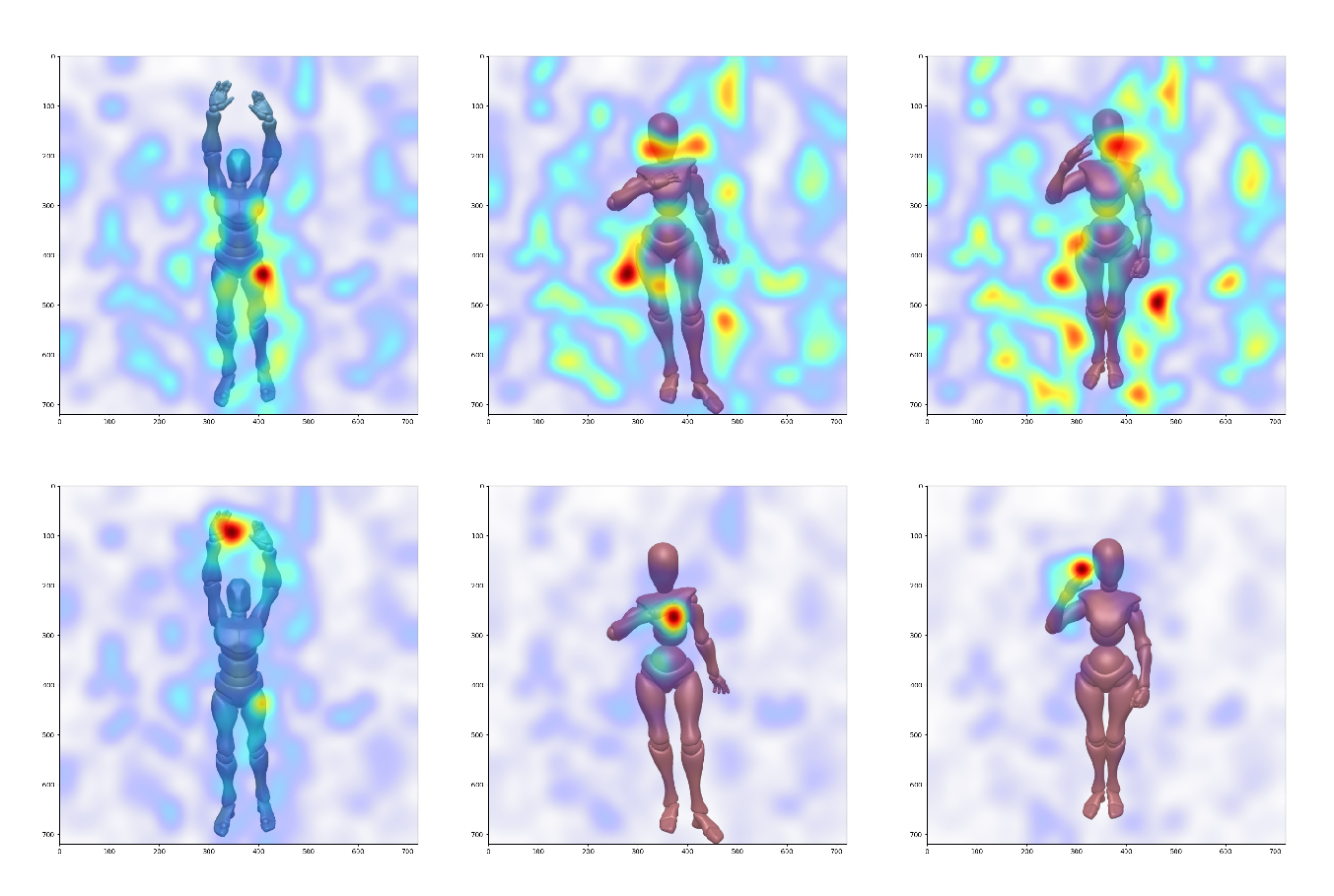}
\caption{
The cross attention map between question and image to generate semantic embedding, the first row is question without prompt and second row is question with prompt. Trough guiding visual question answering, the semantic embedding concentrate on the localized regions which preserve motion semantics.
}
\label{fig:prompt_attention}
\vspace{-0.4cm}
\end{figure}
To gain insights into the preservation of motion semantics, we visualize the attention map between the question and image. In Fig.~\ref{fig:prompt_attention}, we illustrate how the semantic embedding accurately captures motion semantics in localized regions.
This also clarifies the slight changes in the other joints of the character's skeleton with and without semantic consistency.

\subsection{Ablation Study}

\begin{figure}[t]
\centering
\vspace{6pt}
\includegraphics[width=\linewidth]{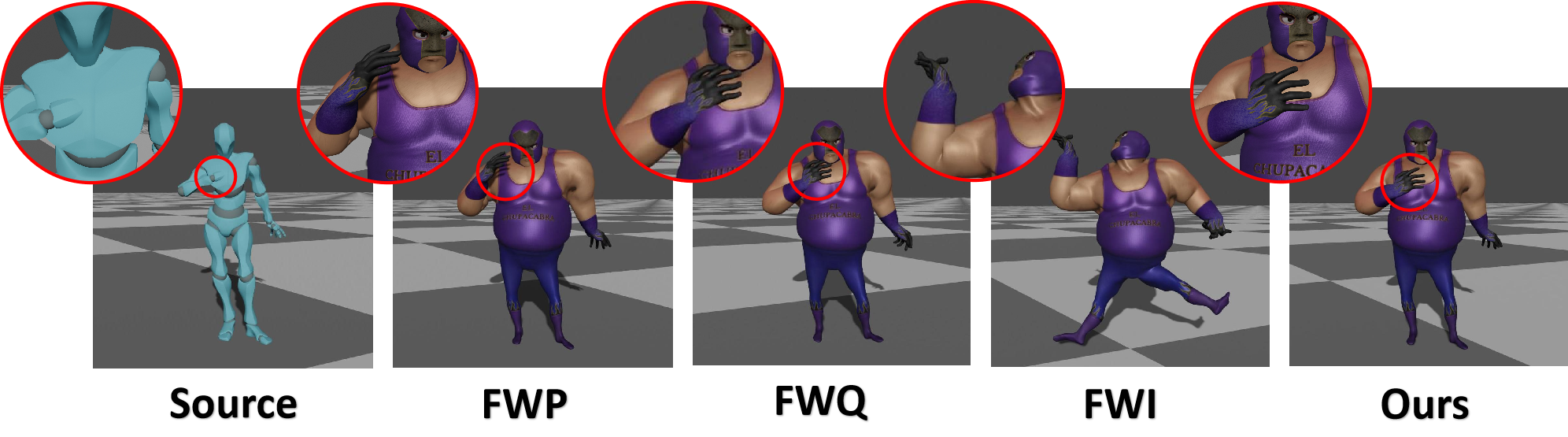}
\caption{
The qualitative comparison between the network fine-tuned without semantics (FWP), the network fine-tuned with the output of querying transformer (FWQ), the network fine-tuned with the output of image encoder (FWI), the network fine-tuned with the output of large language model encoder (Ours).
}
\label{fig:features_ablation}
\vspace{-0.4cm}
\end{figure}

\begin{figure}[t]
\centering
\vspace{6pt}
\includegraphics[width=\linewidth]{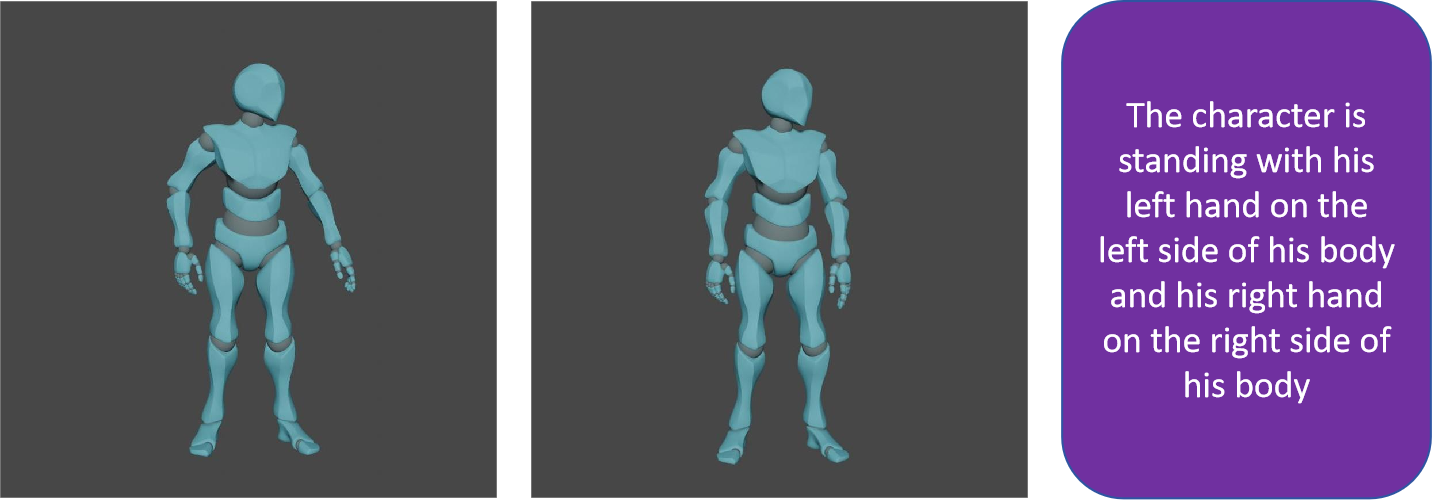}
\caption{
When the motion does not have obvious semantics, the text description can be the same while the semantic embedding from language encoder preserves the difference.
}
\label{fig:fuzzy_text}
\vspace{-0.4cm}
\end{figure}

\begin{figure}[t]
\centering
\includegraphics[width=\linewidth]{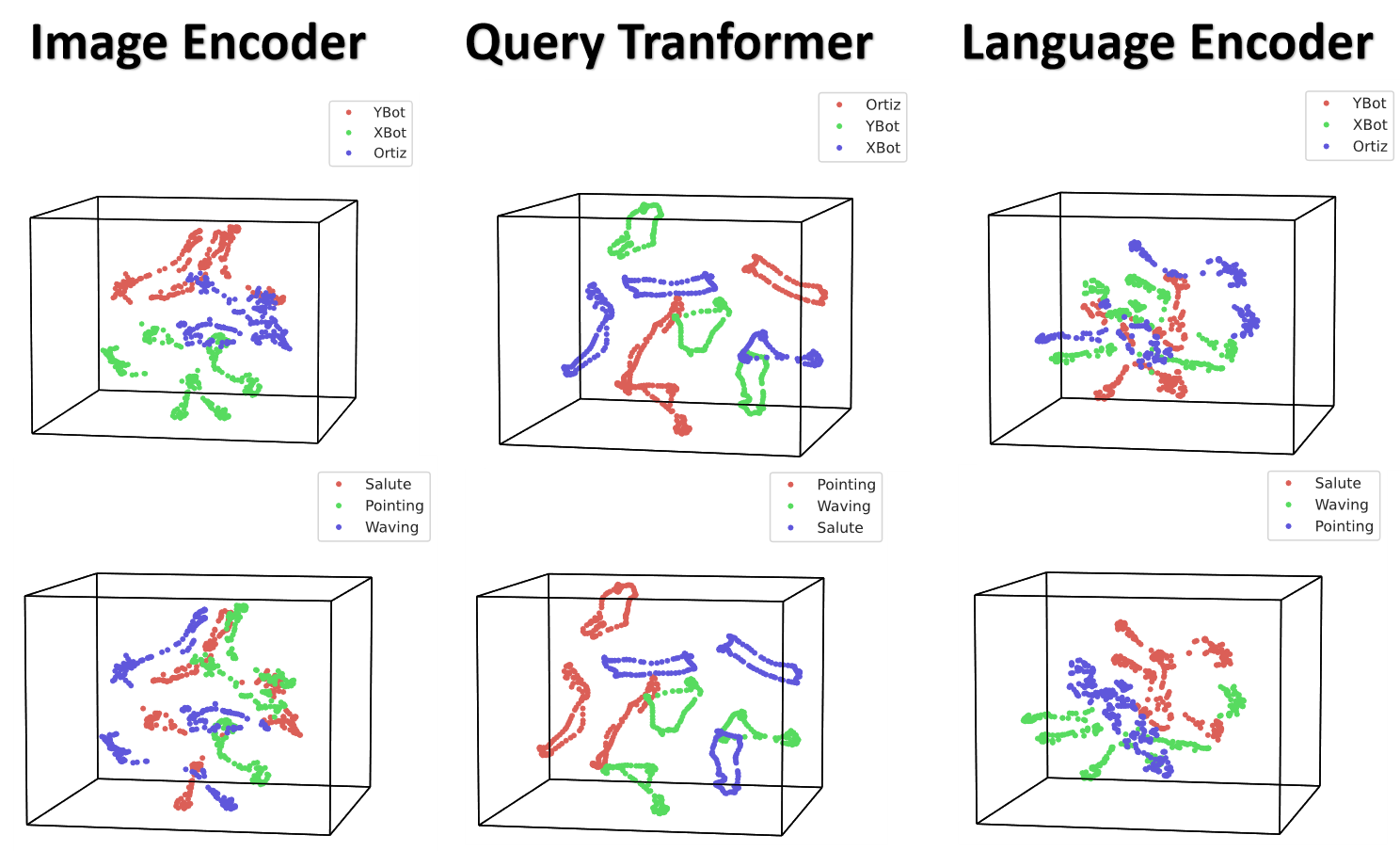}
\caption{
Features extracted from the image encoder, the querying transformer and the encoder of the large language model visualized by T-SNE \cite{van2008visualizing}. 
}
\label{fig:features}
\end{figure}

\begin{figure}
\centering
\includegraphics[width=0.8\linewidth]{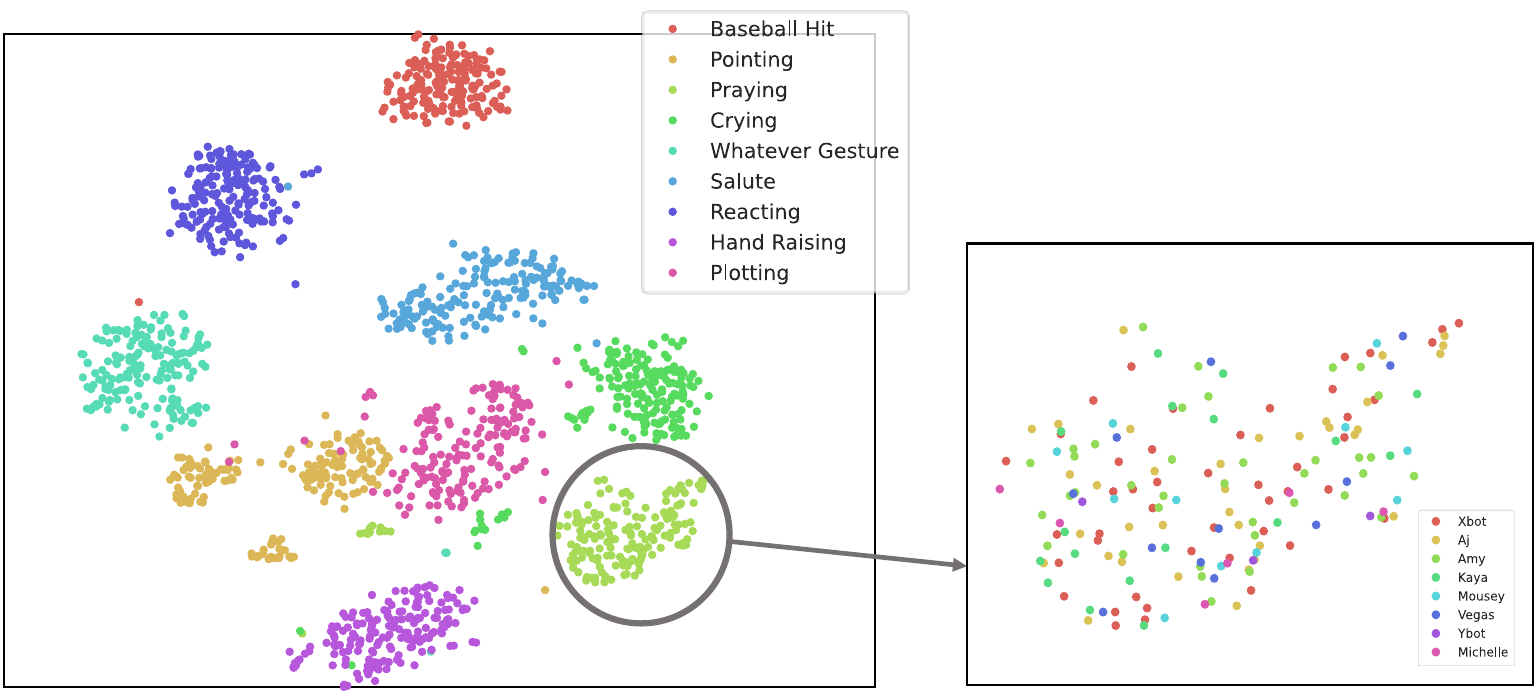}
\caption{
Visualization of semantic embedding space by t-SNE\cite{van2008visualizing}. We visualize 9 motions of 8 characters with 72 sequences in total. 
The left figure demonstrates that frames of the same motion category are clustered. The right one shows that different characters in a single motion category are not clustered by characters.
}
\label{fig:motionvis}
\vspace{-0.6cm}
\end{figure}

\noindent\textbf{Latent semantic embedding.}
We validate features from different levels: the output of the image encoder, the output of the querying transformer, the output of the large language model encoder. We visualize the features of three motions including Waving, Pointing and Salute with three source characters, including Y Bot, X Bot, Ortiz, using T-SNE \cite{van2008visualizing} dimensionality reduction technique. The Fig. \ref{fig:features} shows that the features of the image encoder are clustered around characters rather than motions which indicates that the features contains more information on the appearance of the characters, while the features of the large language model encoder are clustered around motions and contains mainly motion semantics. We further use these features as semantic embedding to fine-tune our model and compute evaluation metrics in Tab.~\ref{tab:compare1}. The metrics and qualitative comparison indicates that appearance of character may lead to meaningless gradient for the model resulting in unnatural retargeted motion. Moreover, compared with the image features from CLIP, the features of the querying transformer and the large language model encoder focus more on relevant semantics with the help of guiding questions.Further more, We visualize the semantic embeddings outputed by VLM in Figure~\ref{fig:motionvis}. The visualization indicates that VLM has captured rich motion semantics information, regardless of the appearance of the character, and is capable of guiding motion semantics preservation.

We also validate the possibility of using text descriptions as semantic embedding. But the text description of motion semantics is fuzzy and sparse. The Fig.~\ref{fig:fuzzy_text} shows that during the transition phase of an action, the text descriptions remain the same. Comparing with text description, the latent features can provide dense supervision. Moreover, using text descriptions or output of decoder will bring more computation cost and higher non-linearity.


\begin{figure*}[t]
\centering
\includegraphics[width=0.6\linewidth]{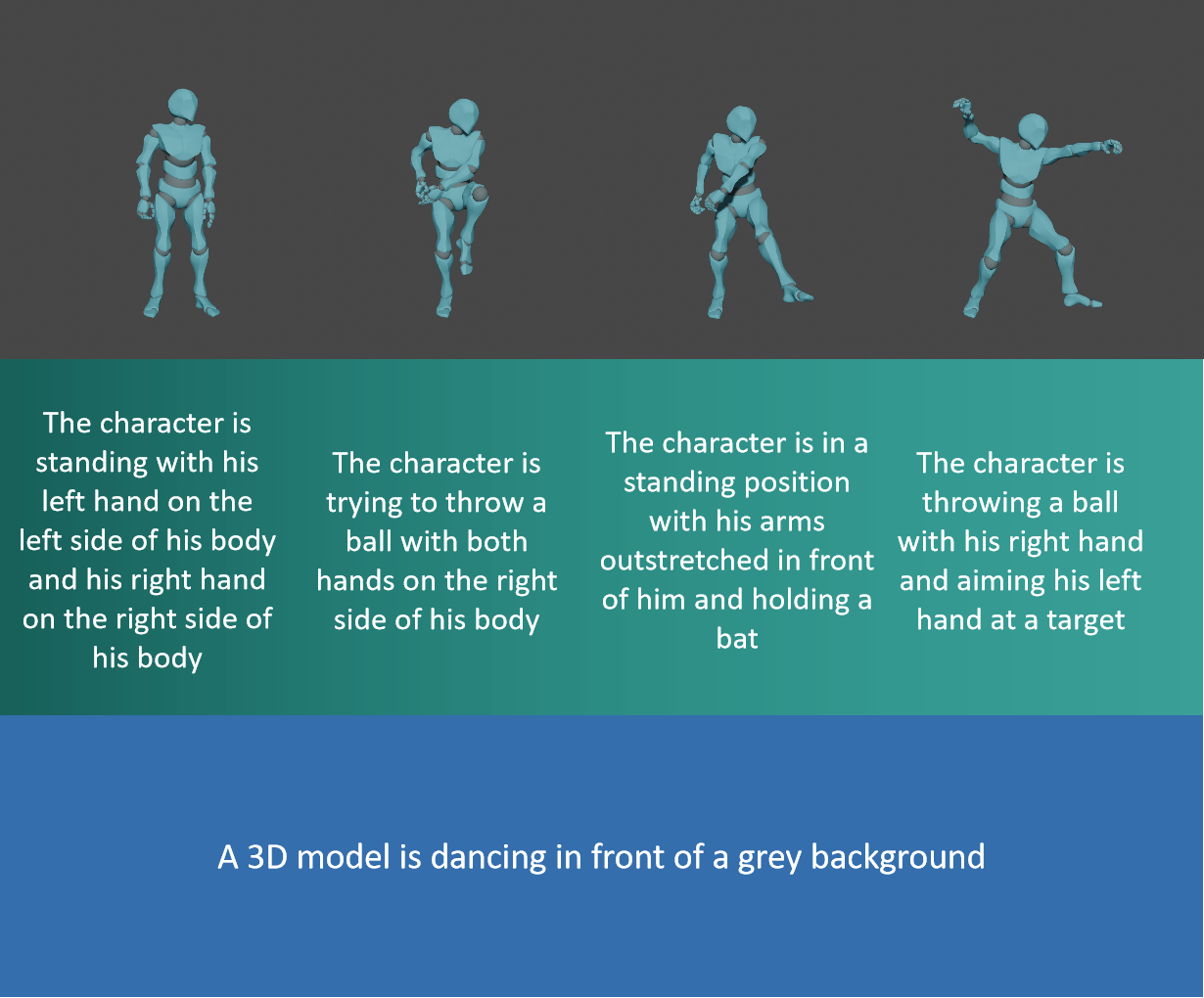}
\caption{
Text descriptions generated by the image language model and video language model. Green row is generated by image-based vision-language model. Blue row is the result of video-based vision-language model.
}
\label{fig:video_image}
\vspace{-0.4cm}
\end{figure*}

\noindent\textbf{Comparison with video semantics.}
We compare the textual descriptions obtained from images and videos of motion sequences. The video-based vision-language model used for evaluation is SwinBert \cite{lin2021end-to-end}. The Fig.~\ref{fig:video_image} shows that although the video-based vision-language model can capture temporal information, the captured semantics is vague and lack details. The image-based vision-language model could generate more detailed and comprehensive descriptions and provide stronger supervision. Moreover, semantics of similar motions could be shared cross different motion sequence, which reduce the size of the fine-tuning set.

\noindent\textbf{Performance of different numbers of views.} We have conducted a new ablation in Table~\ref{tab:view}. The performance decreases to some extent with a single view due to depth information loss. Three perspectives, including the front, left and right view, can achieve fairly good results. Adding additional viewpoints after the third improves the outcome, but at a slower rate.

\noindent\textbf{Ablation study of skeleton network.} We have conducted a new ablation experiment in Table~\ref{tab:backbone}. The results show that the semantic module is applicable for different backbones and improves motion semantics preservation. The MSE metric has increased because the ground truth data in Mixamo dataset are not clean and suffer from interpenetration issues and semantic information loss [27].

\begin{table}
\centering
\resizebox{\columnwidth}{!}{%
\begin{tabular}{@{}ccccccc@{}}
\toprule
Number of Views & MSE $\downarrow$ & MSE$^{lc}$ $\downarrow$ & Pen.\% $\downarrow$ & ITM $\uparrow$ & FID $\downarrow$ &SCL $\downarrow$\\
\midrule
1(front) &0.262  &0.186  &3.51  &0.630  &5.715  &0.499  \\
2(left right diagonal) &0.274  &0.195  &3.49  &0.651  &2.849  &0.277  \\
3(front,left,right) & 0.284 & 0.229 & 3.50 & 0.680 & 0.436 &0.143\\
5(2+3) &0.286  &0.233  &3.50  &0.681  &0.433  &0.141  \\
\bottomrule
\end{tabular}
}
\vspace{-0.3cm}
\caption{Quantitative comparison of different numbers of views.}
\label{tab:view}
\vspace{0.1cm}

\centering
\resizebox{\columnwidth}{!}{%
\begin{tabular}{@{}lcccccc@{}}
\toprule
Method & MSE $\downarrow$ & MSE$^{lc}$ $\downarrow$ & Pen.\% $\downarrow$ & ITM $\uparrow$ & FID $\downarrow$ &SCL $\downarrow$\\
\midrule
NKN [26] & 0.326 & 0.231& 8.71 & 0.575 &27.79 &1.414\\
NKN [26] + VLM &0.392  &0.308   &4.44  &0.665  &2.687  &0.223\\
SAN [2] & 0.435 & 0.255& 9.74 & 0.561 &  28.33& 1.448\\
SAN [2] + VLM &0.481  &0.339    &5.08  &0.659  &2.798  &0.258\\

\bottomrule
\end{tabular}
}
\vspace{-0.3cm}
\caption{Quantitative comparison of different backbone networks.}
\label{tab:backbone}
\vspace{-0.4cm}
\end{table}

\begin{figure}[t]
\centering
\includegraphics[width=\linewidth]{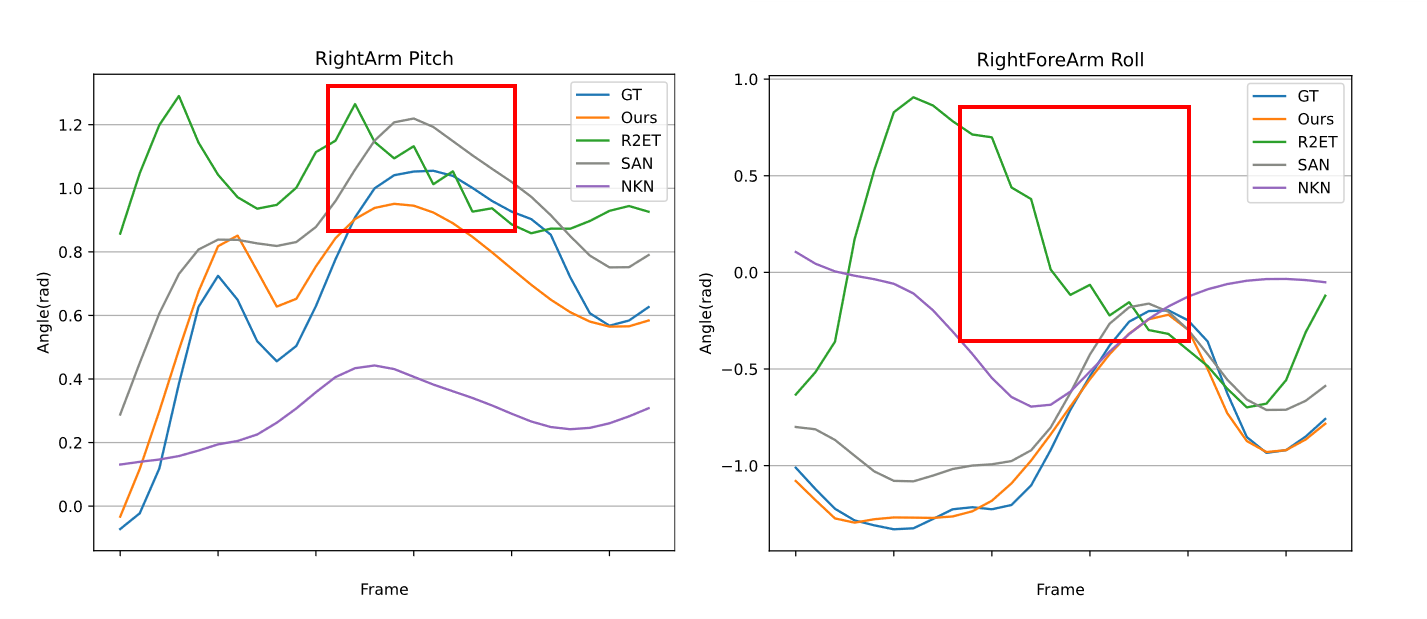}
\caption{Example of the joint angle trajectory for the pitch and roll of the right arm.}
\label{fig:smooth_angle}
\end{figure}

\begin{figure*}[t]
\centering
\includegraphics[width=0.8\linewidth]{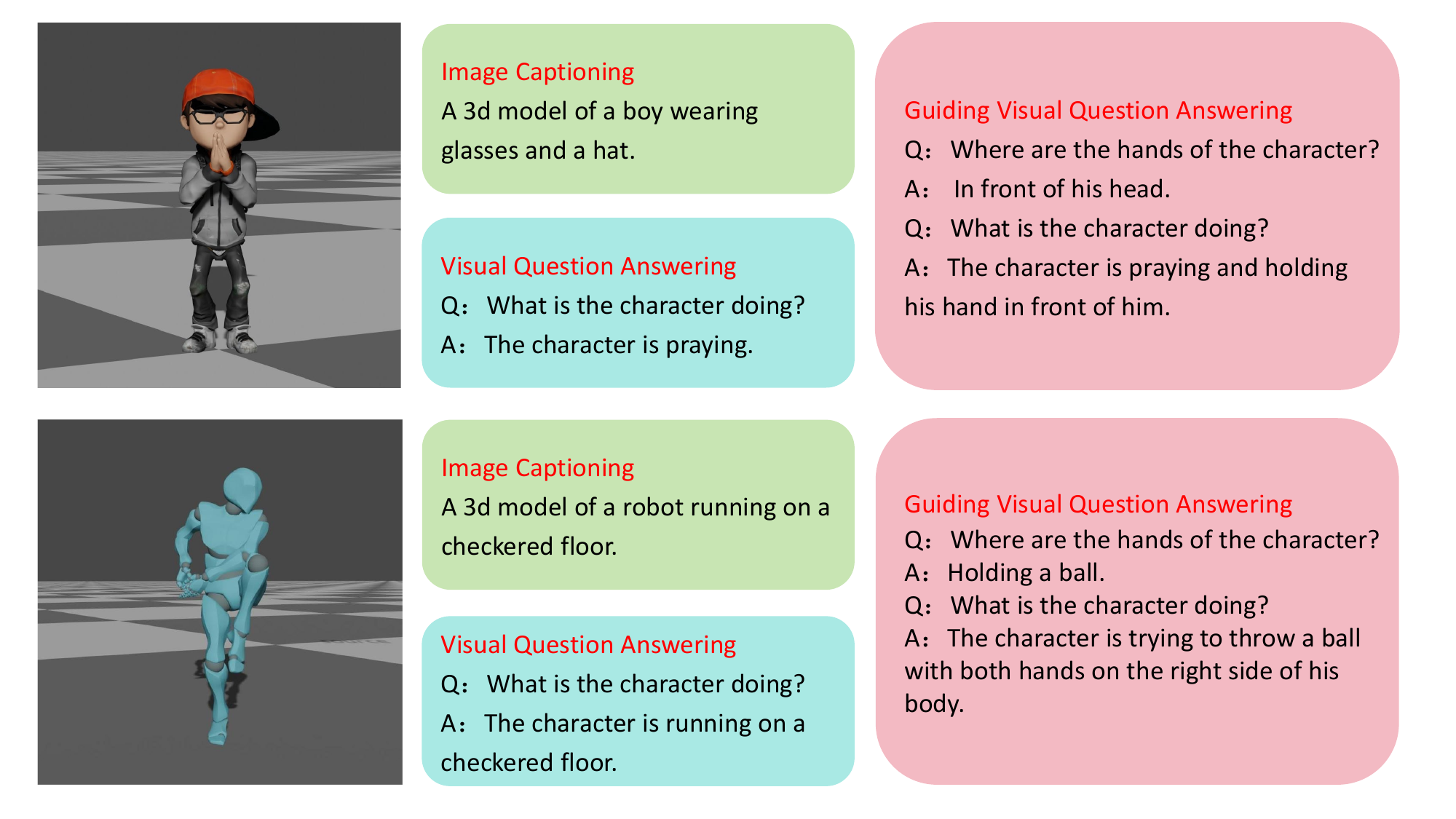}
\caption{
Text descriptions generated by different ways. The guiding visual question answering yields more comprehensive results.
}
\end{figure*}

\subsection{Additional Results}

\begin{figure*}[t]
\centering
\includegraphics[width=0.85\linewidth]{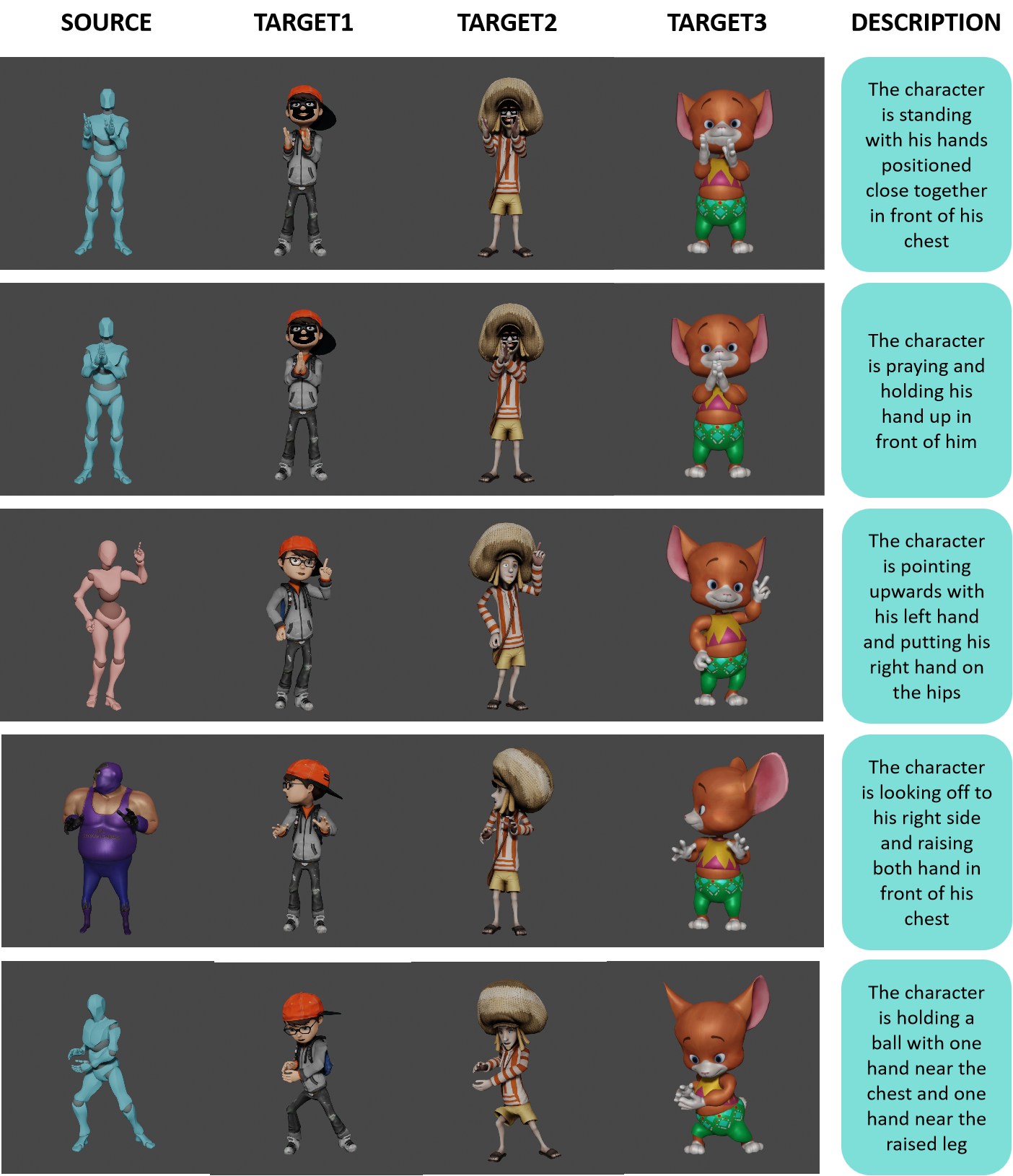}
\caption{Snapshots of motions retargeted from the source character to three different characters and corresponding textual descriptions.}
\label{fig:case_with_text}
\end{figure*}

\begin{figure*}[t]
\centering
\includegraphics[width=0.85\linewidth]{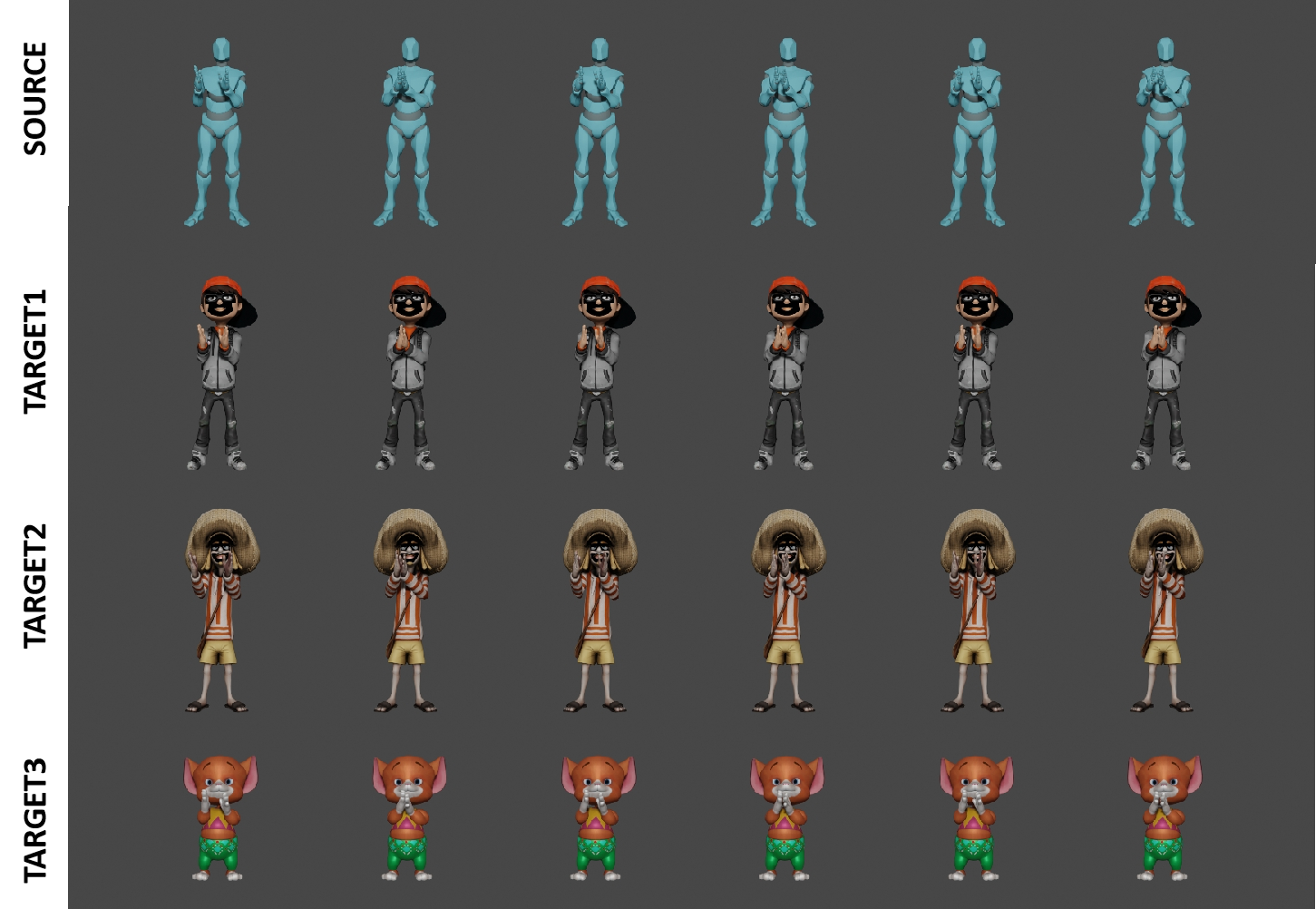}
\caption{Snapshots of motion sequence ``Clapping'' retargeted from the source character to three different characters.}
\label{fig:case1}
\vspace{0.1cm}
\includegraphics[width=0.85\linewidth]{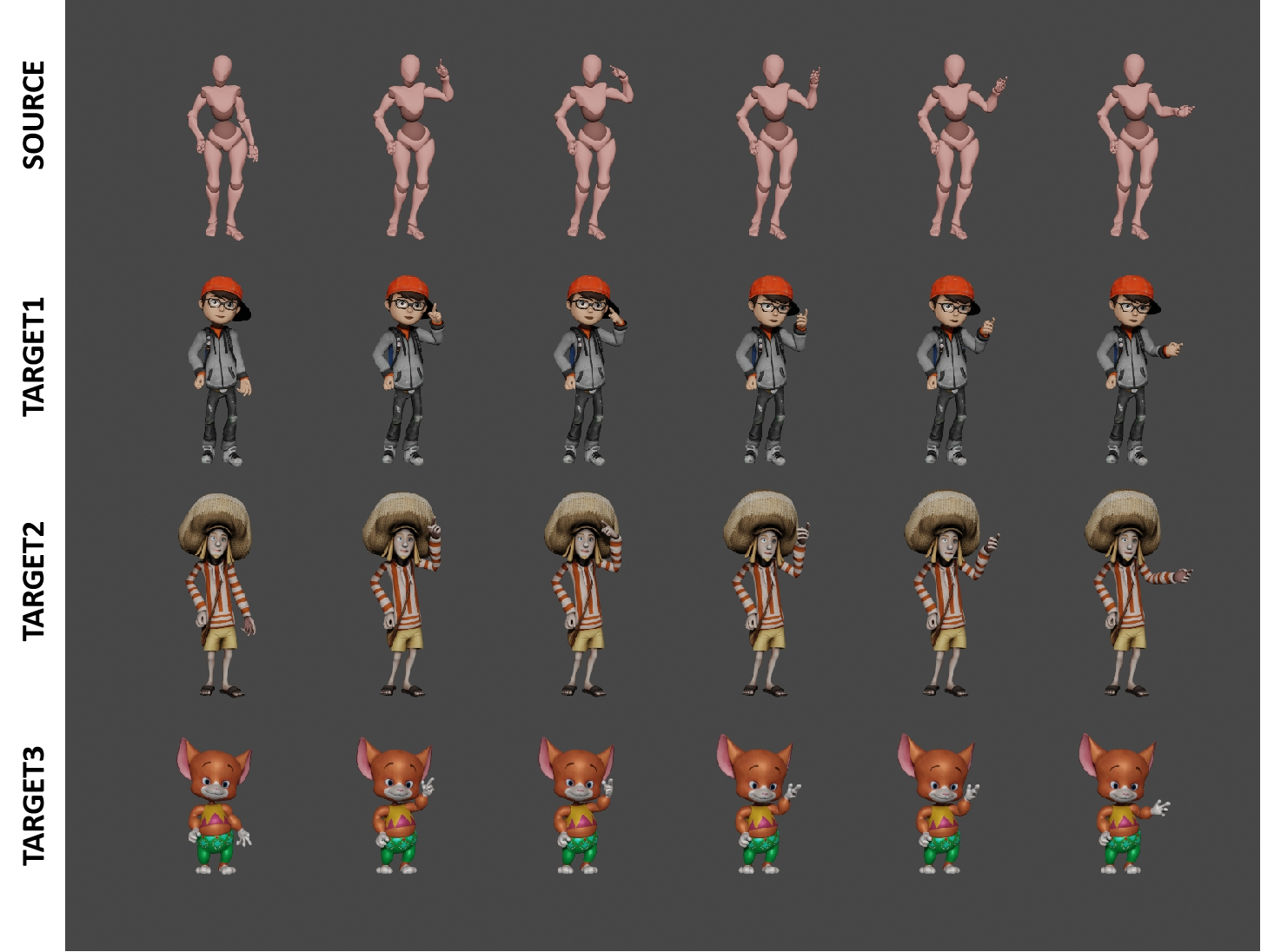}
\caption{Snapshots of motion sequence ``Crazy'' retargeted from the source character to three different characters.}
\label{fig:case2}
\end{figure*}

\begin{figure*}[t]
\centering
\includegraphics[width=0.85\linewidth]{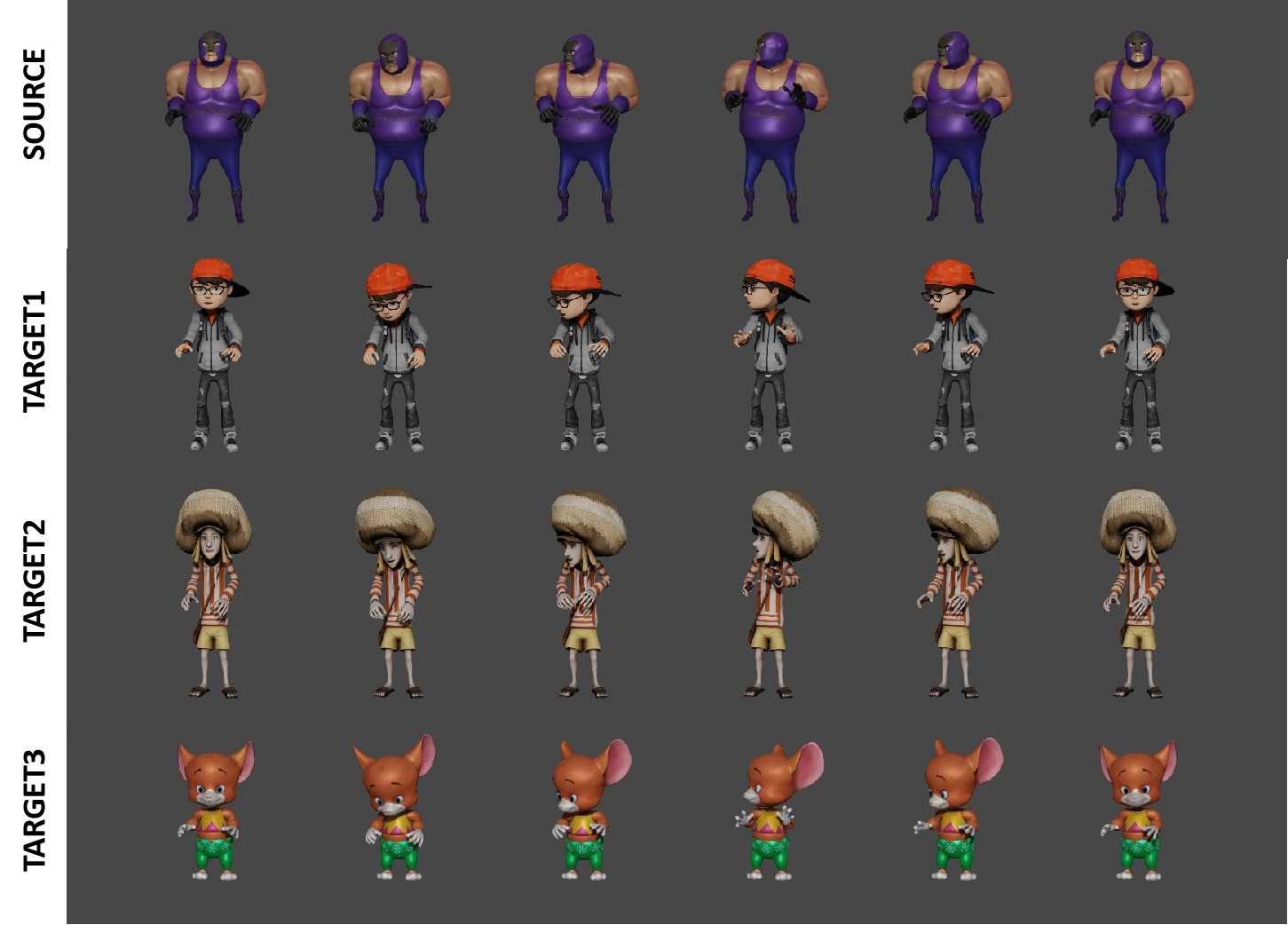}
\caption{Snapshots of motion sequence ``React'' retargeted from the source character to three different characters.}
\label{fig:case3}
\vspace{0.1cm}
\includegraphics[width=0.85\linewidth]{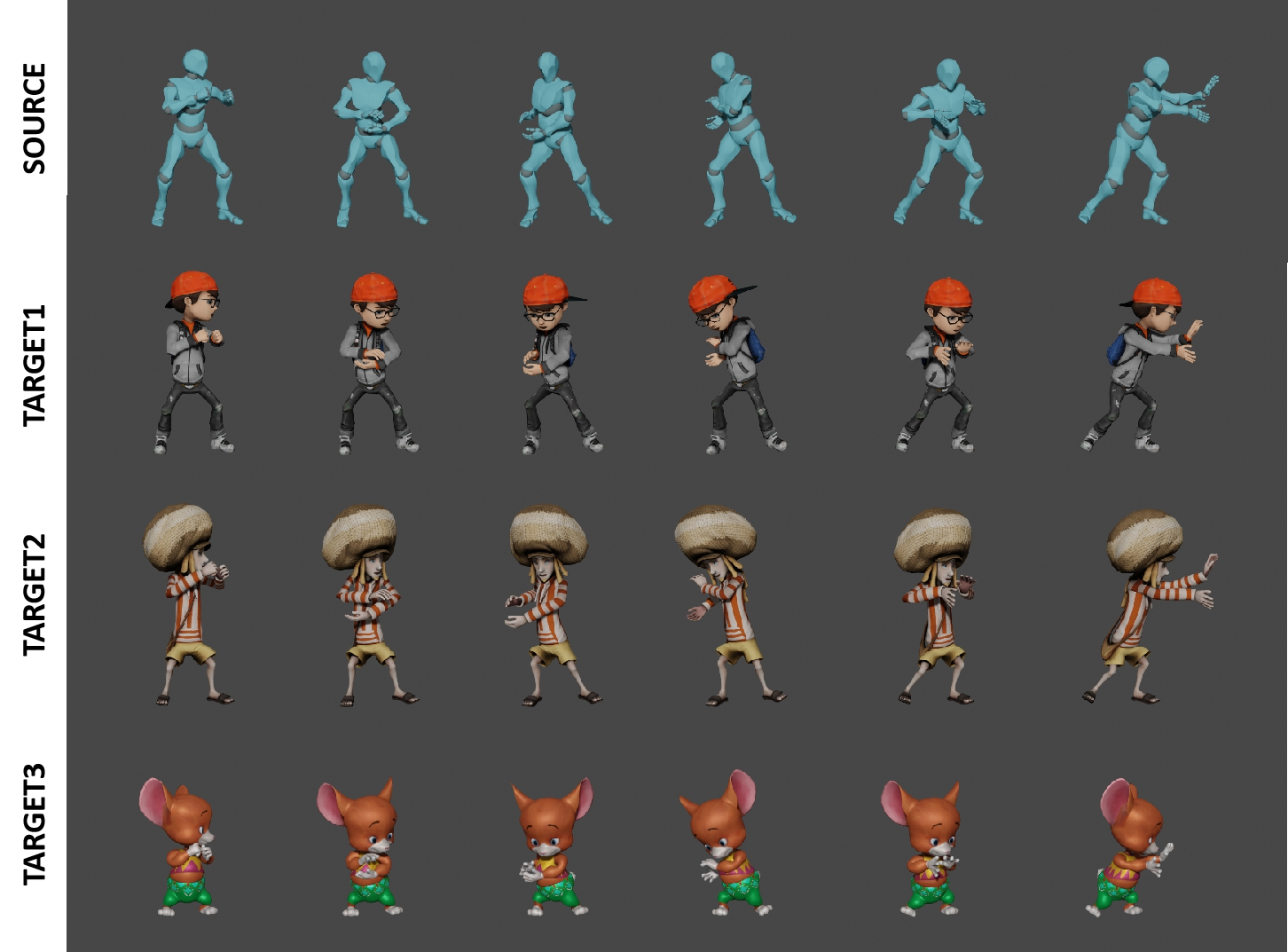}
\caption{Snapshots of motion sequence ``Fireball'' retargeted from the source character to three different characters.}
\label{fig:case4}
\end{figure*}

\noindent\textbf{More cases.} We provide additional cases to validate the effectiveness of the proposed method in the task of semantics-aware motion retargeting. Fig.~\ref{fig:case_with_text} displays a gallery of retargeted results alongside their corresponding textual descriptions. Moreover, Fig.~\ref{fig:case1}, Fig.~\ref{fig:case2}, Fig.~\ref{fig:case3}, and Fig.~\ref{fig:case4} present the retargeted motions of ``Clapping'', ``Crazy'', ``React'', and ``Fireball'' from the source character to three different target characters. These qualitative results demonstrate that our method is able to produce high-quality motion retargeting results while preserving motion semantics.

\begin{figure*}[t]
\centering
\includegraphics[width=0.95\linewidth]{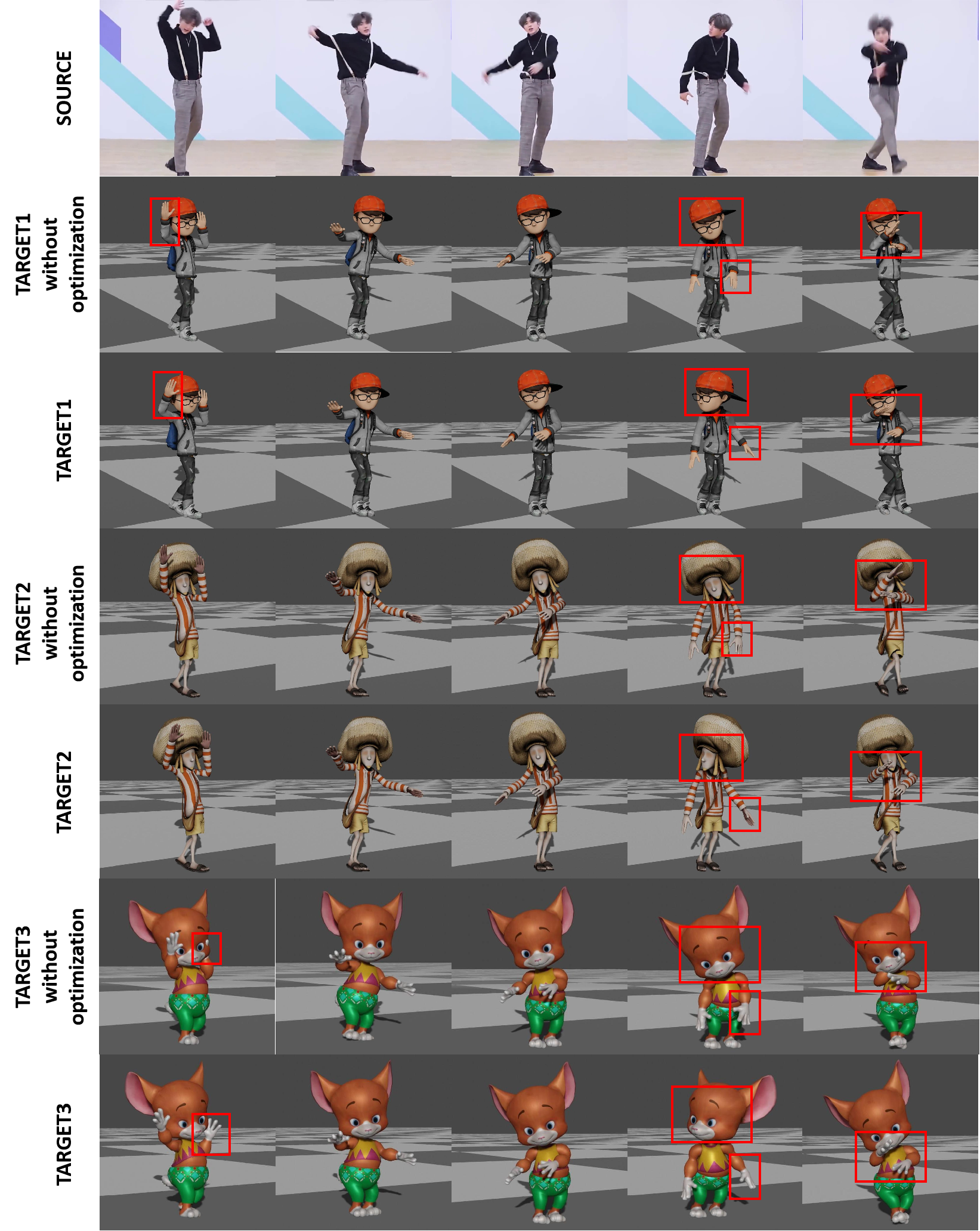}
\caption{Snapshots of motions retargeted from the wild video on the Internet to three different characters with and without optimization.}
\label{fig:case_wild_video}
\end{figure*}

\noindent\textbf{Retargeting motion from Internet videos.} We also conduct experiments on motion retargeting from wild videos on the Internet in Fig.~\ref{fig:case_wild_video}. The human pose is estimated using the approach proposed in \cite{pavllo:videopose3d:2019}.
Then we perform motion retargeting from the human pose to Mixamo characters. Considering the inherent errors in the human pose estimation, we extract the semantic embedding from the original video and apply the semantics consistency loss to further optimize the joint angles.
We compare the results with and without optimization to validate the effectiveness of the semantics consistency loss.
Despite the presence of some errors in human pose estimation, the retargeted motion accurately preserves the motion characteristics of the movement in the original video.

\noindent\textbf{Smoothness.}
We perform experiments to evaluate the smoothness of the retargeted motions. As an example, we visualize the joint angle trajectory for the pitch and roll of the right arm, and compare it with state-of-the-art methods. The Fig.~\ref{fig:smooth_angle} illustrates that our method delivers smoother motion compared to R2ET \cite{zhang2023skinned}.

\begin{table*}[ht]
  \centering
  \begin{tabular}{@{}llllll@{}}
    \toprule
    Index & Motion name & Search query& Source character & Length & Usage\\
    \midrule
    1 & Agreeing & Step Back Cautiously Agreeing & Y Bot & 142 & Test \\
    2 & Angry & Standing Angrily & Y Bot & 576 & Finetune \\
    3 & Baseball Hit & Baseball Base Hit & X Bot & 118 & Test \\
    4 & Baseball Pitching & Pitching A Baseball & Y Bot & 119 & Test \\
    5 & Cards & Dealing Cards & X Bot & 274 & Finetune \\
    6 & Charge& Point Onward Charge& Y Bot& 172&Test\\
    7 & Clapping & Clap While Standing & Y Bot & 36 & Test \\
    8 &Counting& Counting To Five On One Hand& Y Bot& 200& Test\\
     9& Crying & Crying And Rubbing Eyes & X Bot & 189 & Test \\
    10& Crazy Gesture & Crazy Hand Gesture & X Bot &151&Test\\ 
     11& Defeat & Covering Face In Shame After Defeat & Y Bot & 220 & Finetune\\
     12& Dismissing Gesture& Dismissing With Hand Forward& Y Bot& 99&Test\\
     13& Excited & Super Excited & X Bot& 198 & Finetune \\
     14& Fireball & Street Fighter Hadouken & Y Bot & 102 & Test \\
     15& Fist Pump & Pymping A Fist & Y Bot & 115 & Finetune \\
    16&Focus & Shake Off Head Pain And Focus & X Bot & 166 & Finetune \\
    17&Guitar Playing & Playing A Guitar & Y Bot & 144 & Test \\
    18&Happy& Standing Happily & X Bot & 301 & Test \\
    19&Hands Forward Gesture& Two Handed Forward Gesture& Ortiz&94&Test\\
    20&Hand Raising& Raising A Hand& X Bot& 123& Test\\
    21&Insult & Insulting With Rude Gesture& X Bot&81& Test\\
    22&Lead Jab & Long Body Jab & Y Bot & 56 & Test \\
    23&Looking&Looking Off Into The Distance& Y Bot& 241& Test\\
    24&Loser& Showing Loser Gesture While Standing& Ortiz& 99& Test\\
    25&No & Indicating No & X Bot & 151 & Test \\
    26&Padding & Padding A Single Oar Canoe& Y Bot & 218 & Finetune \\
    27&Plotting& Evil Plotting& X Bot& 100& Test\\
    28&Pointing& Pointing While Seated& Ortiz& 104 & Test\\
    29&Praying & Buckled Stand And Praying & Y Bot & 36 & Test \\
    30& Reacting& Being Surprised And Looking Right& Oritz& 111& Test\\
    31&Salute & Formal Military Salute & X Bot & 86 & Test \\
    32&Shaking Hands 2& 2 People Shaking Hands Part 2 - Male& Y Bot& 132& Test\\
    33&Smoking & Idle Smoking & Y Bot & 538 & Test \\
    34&Standing Greeting & Greeting While Standing & Ortiz & 154 & Test \\
    35&Thankful & Being Thankful While Standing & X Bot & 91 & Test \\
    36&Taunt & Taunting Pointing At Wrist& X Bot & 86 & Test \\
    37&Taunt Gesture&Taunt Gesture & Ortiz & 60 & Finetune \\
    38&Talking & Asking A Question & X Bot & 156 & Finetune \\
    39&Talking & Male Talking On The Cell Phone & Y Bot & 145& Finetune \\
    40&Telling A Secret & Telling A Secret& Ortiz & 328 & Finetune \\
    41&Victory& Celebrating After A Win While Seated & Ortiz & 184 & Finetune \\
    42&Waving & Waving With Both Hands & Ortiz & 96 & Finetune \\
    43&Whatever Gesture & Whatever Gesture& Ortiz & 46 & Test \\ 
    44&Yawn & Big Yawn While Standing & X Bot & 251 & Finetune \\
    45&Yelling & Yelling In Anger& Ortiz & 236 & Finetune \\
    \bottomrule
  \end{tabular}
  \caption{45 source motion sequences of 3 characters for testing and fine-tuning.}
  \label{tab:dataset}
\end{table*}

\end{document}